\pdfoutput=1
\documentclass[mnsc,nonblindrev]{informs3_hide}

\usepackage{xcolor}

\definecolor{DarkBlue}{rgb}{0.0,0.0,0.55}
\usepackage[backref = false, bookmarks, colorlinks = true, plainpages = false, citecolor = DarkBlue , urlcolor = DarkBlue, filecolor = DarkBlue, linkcolor = DarkBlue]{hyperref}

\usepackage{algorithm}
\usepackage[noend]{algorithmic}
\usepackage{booktabs}
\usepackage{diagbox}
\usepackage{appendix}
\usepackage{hhline}
\usepackage{bm}
\usepackage[normalem]{ulem}
\usepackage{multirow,multicol,xspace,enumitem,subcaption,caption}
\usepackage[flushleft]{threeparttable}

\newcommand{\cD}{{\mathcal{D}_{\mathcal{X}}}}
\newcommand{\mcD}{{\mathcal{D}}}
\newcommand{\cF}{{\mathcal{F}}}
\newcommand{\cA}{{\mathcal{A}}}
\newcommand{\cX}{{\mathcal{X}}}
\newcommand{\cE}{{\mathcal{E}}}
\newcommand{\cV}{{\mathcal{V}}}
\newcommand{\cR}{{\mathcal{R}}}
\newcommand{\wcR}{\widehat{\mathcal{R}}}
\newcommand{\Reg}{{\mathrm{Reg}}}
\newcommand{\wReg}{\widehat{\mathrm{Reg}}}
\newcommand{\cB}{\mathcal{B}}
\newcommand{\cP}{\mathcal{P}}

\newcommand{\E}{{\mathbb{E}}}

\newcommand{\1}{\mathbb{I}}
\newcommand{\N}{\mathbb{N}}

\newtheorem{observation}{Observation}
\newtheorem{setup}{Setup}

\usepackage{natbib}
\bibpunct[, ]{(}{)}{,}{a}{}{,}%
\def\bibfont{\small}%

\TheoremsNumberedThrough     %
\ECRepeatTheorems

\EquationsNumberedThrough    %

\MANUSCRIPTNO{}

\begin{document}

\TITLE{\Large{Bypassing the Monster: A Faster and Simpler Optimal Algorithm for Contextual Bandits under Realizability}}

\ARTICLEAUTHORS{
\AUTHOR{David Simchi-Levi ~~~~~~~~~~ Yunzong Xu ~~~~}
\AFF{Institute for Data, Systems, and Society, Massachusetts Institute of Technology\\dslevi@mit.edu, yxu@mit.edu}
}

\ABSTRACT{
We consider the general (stochastic) contextual bandit problem under the realizability assumption, i.e., the expected reward, as a function of contexts and actions, belongs to a general  function class $\cF$. We design a fast and simple algorithm that achieves the statistically optimal regret with only ${O}(\log T)$ calls to an offline  regression oracle across all $T$ rounds. The number of oracle calls can be further reduced to $O(\log\log T)$ if $T$ is known in advance.  Our results provide the first universal and optimal reduction from contextual bandits to offline regression, solving an important open problem in the contextual bandit literature. A direct consequence of our results is that any advances in offline regression immediately translate to contextual bandits, statistically and computationally. This leads to faster algorithms and improved regret guarantees for broader classes of contextual bandit problems. 
}

\HISTORY{\text{Submitted} April 2020; \text{Revised} April 2021; \text{Accepted} June 2021, by \text{Mathematics of Operations Research}}

\maketitle

\section{Introduction}
The contextual bandit problem is a fundamental framework for online decision making and interactive machine learning, with diverse applications ranging from healthcare (\citealt{tewari2017ads,bastani2020online}) to electronic commerce (\citealt{li2010contextual,agarwal2016making}).  It has been extensively studied in  computer science, operations research, and statistics literature.

Broadly speaking, approaches to contextual bandits can be classified into two categories (see \citealt{foster2018practical}): \textit{realizability-based} approaches which rely on weak or strong assumptions on the model representation, and \textit{agnostic} approaches which are completely model-free. While many different contextual bandit algorithms (realizability-based or agnostic) have been proposed over the past twenty years, most of them suffer from either theoretical or practical issues (see \citealt{bietti2018contextual}). Existing \text{realizability-based} algorithms  building on upper confidence bounds (e.g., \citealt{filippi2010parametric,abbasi2011improved,chu2011contextual,li2017provably}) and Thompson sampling (e.g., \citealt{agrawal2013thompson,russo2018tutorial}) rely on strong assumptions on the model representation and are only tractable for specific parametrized families of models like generalized linear models. Meanwhile,  \text{agnostic} algorithms that make no assumption on the model representation (e.g., \citealt{dudik2011efficient,agarwal2014taming}) may lead to overly conservative exploration in practice (\citealt{bietti2018contextual}), and their reliance on an \textit{offline cost-sensitive classification oracle} as a subroutine typically causes implementation difficulties  as the oracle itself is computationally intractable in general. At this moment, designing a provably optimal contextual bandit algorithm that is applicable for large-scale real-world deployments is still widely deemed a very challenging task (see \citealt{agarwal2016making,foster2020beyond}).

Recently, \cite{foster2018practical} propose an approach to solve contextual bandits with general model representations (i.e., general function classes) using an \textit{offline regression  oracle} --- an oracle that can typically be implemented efficiently and has wide availability for numerous function classes due to its core role in modern machine learning. %
{Specifically, motivated by the  work of  \cite{krishnamurthy2019active} which initiates such a key idea, \cite{foster2018practical} assume access to a \emph{weighted least squares regression oracle}, which is deemed highly practical as it has a strongly convex loss function and is amenable to gradient-based methods.} 
As \cite{foster2018practical} point out, designing offline-regression-oracle-based algorithms is a promising direction for making contextual bandits practical, as they seem to combine the advantages of both realizability-based and agnostic algorithms: they are general and flexible enough to work with any given function class, while using a more realistic and reasonable oracle than the computationally-expensive classification oracle. Indeed, according to multiple experiments and extensive empirical evaluations conducted by  \cite{foster2018practical} and \cite{bietti2018contextual}, the algorithm of \cite{foster2018practical} ``works the best overall'' among existing contextual bandit approaches.

Despite its empirical success, the algorithm of \cite{foster2018practical} is, however, theoretically sub-optimal --- it could incur $\widetilde{\Omega}(T)$ regret in the worst case. Whether the optimal regret of contextual bandits can be attained via an offline-regression-oracle-based algorithm is listed as an open problem in \cite{foster2018practical}. In fact, this problem has been open to the bandit community since 2012 --- it dates back to  \cite{agarwal2012contextual}, where the authors propose a computationally \textit{inefficient} contextual bandit algorithm that achieves the optimal $\widetilde{O}(\sqrt{KT\log|\cF|})$ regret for a general \textit{finite} function class $\cF$, but leave designing computationally tractable algorithms as an open problem.

More recently, \cite{foster2020beyond} propose an algorithm that achieves the optimal regret for contextual bandits by assuming access to an \textit{online regression oracle} (which is not an {offline  oracle} and has to work with an adaptive adversary). Their finding that contextual bandits can be reduced to online regression is novel and important, and their result is also very general: it requires only the minimal realizability assumption, and holds true even when the contexts are chosen adversarially. However, compared with access to an offline regression oracle, access to an online regression oracle is a much stronger (and relatively restrictive) assumption.  In particular, optimal and efficient algorithms for online regression are only known for specific function classes, much less than those known for offline regression. Whether the optimal regret of contextual bandits can be attained via a reduction to an offline regression oracle is listed as an open problem again in \cite{foster2020beyond}.

\subsection{Our Contributions}

In this paper, we study the following question repeatedly mentioned in the contextual bandit literature (\citealt{agarwal2012contextual,foster2018practical,foster2020beyond}): \textit{Is there an offline-regression-oracle-based algorithm that achieves the optimal regret for general (stochastic) contextual bandits?}

We answer this question in the affirmative by providing the first optimal black-box reduction from contextual bandits to offline regression, with only the minimal realizability assumption. The significance of this result is that it reduces contextual bandits, a prominent online decision-making problem,
to offline regression, a very basic and common supervised learning task that serves as the building block of modern machine learning. A consequence of this result is that any
advances in solving offline regression problems translate to contextual bandits, statistically and computationally. Note that such online-to-offline reductions are highly nontrivial for online learning problems in general; in fact, a generic reduction from fully adversarial online learning to offline learning is not possible (\citealt{hazan2016computational}).

Our reduction is accomplished by providing a surprisingly fast and simple algorithm (which builds on and connects the approaches of  \citealt{abe1999associative,agarwal2014taming,foster2020beyond}) and proving strong theoretical guarantees for this algorithm. 
For a general finite function class $\cF$, our algorithm achieves the optimal $\widetilde{O}(\sqrt{KT\log|\cF|})$ regret  %
with only $O(\log T)$ calls to an offline least squares regression oracle over $T$ rounds. The number of oracle calls can be further reduced to $O(\log\log T)$ if $T$ is known. Notably, this can be understood as a ``triply exponential'' improvement over previous work: (i) compared with the previously known regret-optimal algorithm of \cite{agarwal2012contextual} for this setting, which requires enumerating over $\cF$ at each round, our algorithm accesses the function class only through a least squares regression oracle, thus typically avoids an exponential computational cost at each round; (ii) compared with %
the classification-oracle-based algorithm of \cite{agarwal2014taming} which requires $\widetilde{O}(\sqrt{KT/\log|\cF|})$ calls to a computationally expensive classification oracle, our algorithm requires only $O(\log T)$ calls to a simple regression oracle, which implies an exponential improvement (in the number of oracle calls) over existing provably optimal oracle-efficient algorithms, even when we ignore the difference between regression and classification oracles; (iii) when the number of rounds $T$ is known in advance, our algorithm can further reduce the number of oracle calls to $O(\log\log T)$, which is an exponential improvement by itself. Our algorithm is thus highly practical; see Table \ref{tab} for a detailed comparison with existing work. %

\begin{table}[htbp]
\renewcommand*{\arraystretch}{1.2}
\centering
\caption{Algorithms' performance with general finite $\cF$ and i.i.d. contexts.  Advantages are marked in bold.}\label{tab}
\begin{threeparttable}
\begin{tabular}{ |p{4.5cm}||p{3cm}|p{6.5cm}|  }
 \hline
 Algorithm  & Regret rate & Computational complexity\\
\hhline{|-|-|-|}
\hline
\texttt{Regressor Elimination} & \textbf{optimal}  &$\Omega(|\cF|)$\\
 (\citealt{agarwal2012contextual})   & & {intractable}  \\
 \hline
 \texttt{ILOVETOCONBANDITS} & \textbf{optimal} & $\widetilde{O}(\sqrt{KT/\log|\cF|})$ calls to an\\
(\citealt{agarwal2014taming}) &  &  offline {classification} oracle\\
\hline
\texttt{RegCB}& {suboptimal} &$O(T^{3/2})$ calls to  an\\
 (\citealt{foster2018practical}) &  &   offline least squares oracle\\
 \hline
 \texttt{SquareCB} & \textbf{optimal} & $O(T)$ calls to an \\
 (\citealt{foster2020beyond})    &  & {online} regression oracle  \\
 \hline
 \texttt{FALCON} / \texttt{FALCON+} & \textbf{optimal} & $\bm{O(\log T)}$ or $\bm{O(\log{} \log T)}$ calls to an\\
 (this paper) &   &  \textbf{offline regression} oracle* \\
 \hline
\end{tabular}
\begin{tablenotes}\renewcommand*{\arraystretch}{1}
\small \item[*] Not restricted to least squares; strictly easier to solve than online regression. See \S\ref{sec:extension} for details.
\end{tablenotes}
\end{threeparttable}
\end{table}

We  then extend all of the above results to the general setting where (i) the function class $\cF$ can be infinite, and (ii) the offline regression oracle is not necessarily a least squares oracle. For this general setting, our reduction can be stated as follows: 
for any function class $\cF$, given an arbitrary offline regression oracle with an arbitrary offline \textit{estimation error} (or \emph{excess risk}) guarantee, we provide a fast and simple contextual bandit algorithm whose regret can be bounded by a function of the offline  estimation error, through only $O(\log T)$  calls (or $O(\log\log T)$ calls if $T$ is known) to the offline regression oracle. %
We show that our algorithm is statistically optimal as long as the offline regression oracle is statistically optimal. Notably, the above results provide a universal and optimal  ``converter'' from  results of offline regression with general function classes to results of contextual bandits with general function classes. This leads to improved algorithms with tighter regret bounds for many existing contextual bandit problems, as well as practical algorithms for many new contextual bandit problems, e.g., contextual bandits with certain types of neural networks, and contextual bandits with heavy-tailed rewards.

The analysis of our algorithm is particularly interesting. Unlike existing analysis of other realizability-based algorithms in the literature, we do not directly analyze the decision outcomes of our algorithm ---  instead, we find a dual interpretation of our algorithm as sequentially maintaining a \textit{dense} distribution over \textit{all} (possibly \textit{improper}) policies, where a policy is defined as a deterministic decision function mapping contexts to actions. We analyze how the realizability assumption enables us to establish uniform-convergence-type results for some \textit{implicit} quantities in the universal policy space, regardless of the huge capacity of the universal policy space. Note that while the dual interpretation itself is not easy to compute in the universal policy space, it is only applied for the purpose of analysis and has nothing to do with our original algorithm's implementation. Through this lens, we find that our algorithm's dual interpretation
satisfies a series of sufficient conditions for optimal contextual bandit learning.  Our identified sufficient conditions for optimal contextual bandit learning in the universal policy space build on the previous work of \cite{dudik2011efficient}, \cite{agarwal2012contextual} and \cite{agarwal2014taming} ---  the first one is colloquially referred to as the ``monster paper'' by its authors due to its complexity, and the third one is titled as ``taming the monster'' by its authors due to its improved computational efficiency. Since our algorithm achieves all the conditions required for regret optimality in the universal policy space in a completely \textit{implicit} way (which means that all the requirements are automatically satisfied without explicit computation), our algorithm comes with significantly reduced computational cost  compared with previous work (thanks to the realizability assumption), and we thus title our paper as ``bypassing the monster.''

Overall, our algorithm is fast and simple, and our analysis is quite general. We believe that the algorithm has the potential to be implemented on a large scale, and our analysis may contribute to deeper understanding of contextual bandits.
We will go over the details in the rest of this article.

\subsection{Learning Model}\label{sec:setting}
The general stochastic contextual bandit problem can be stated as follows.  Let $\cA$ be a finite set of $K$ actions and $\cX$ be an arbitrary space of contexts (e.g., a feature space). The interaction between the learner and nature happens over $T$ rounds, where $T$ is possibly unknown. At each round $t$, nature samples a context $x_t\in\cX$ and a context-dependent reward vector $r_t\in[0,1]^{\cA}$ according to a fixed but unknown (joint) distribution $\mcD$, with component $r_t(a)$ denoting the reward for action $a\in\cA$; 
the learner observes $x_t$, picks an action $a_t\in\cA$, and observes the reward for her action $r_t(a_t)$. Notably, the learner's reward $r_t(a_t)$ depends on {both} the context $x_t$ and her action $a_t$, and is a {partial observation} of the full reward vector $r_t$. Depending on whether there is an assumption about nature's reward model, prior literature studies the contextual bandit problem in two different but closely related settings.

\textbf{Agnostic setting.} 
Let $\Pi\subset\cA^{\cX}$ be a class of \textit{policies} (i.e., decision functions) that map contexts $x\in\cX$ to actions $a\in\cA$, and $\pi_*=\argmax_{\pi\in\Pi}\E_{(x,r)\sim\mcD}[r(\pi(x))]$ be the optimal policy in $\Pi$ that maximizes the expected reward. The learner's goal is to compete with the (in-class) optimal policy $\pi_*$ and minimize her (empirical cumulative) \textit{regret} after $T$ rounds, which is defined as 
$$
 \sum_{t=1}^T(r_t(\pi_*(x_t))-r_t(a_t)).
$$
The above setting is called \textit{agnostic} in  the sense that it imposes no assumption on nature.

\textbf{Realizable setting.} 
Let $\cF$ be a class of \textit{predictors} (i.e., reward functions), where each predictor is a function $f:\cX\times\cA\rightarrow[0,1]$ trying to approximate the \emph{true reward function} $f^*$ defined by $f^*(x,a)=\E[r_t(a)\mid x_t=x],~\forall x\in\cX, a\in\cA$. The standard realizability assumption  (\citealt{chu2011contextual,agarwal2012contextual,foster2018practical}) is as follows:
\begin{assumption}[Realizability] The true reward function is contained in $\cF$, i.e., $f^*\in\cF$.
\end{assumption}
Given a predictor $f\in\cF$, the associated reward-maximizing policy $\pi_{f}$ always picks the action with the highest predicted reward, i.e., $\pi_{f}(x)=\arg\max_{a\in\cA}f(x,a)$. The learner's goal is to compete with the \emph{globally optimal policy} $\pi_{f^*}$ and minimizes her (empirical cumulative) \textit{regret} after $T$ rounds, which is defined as 
$$
 \sum_{t=1}^T(r_t(\pi_{f^*}(x_t))-r_t(a_t)).
$$
The above setting is called \textit{realizable} in the sense that it assumes that nature can be well-specified by a predictor in $\cF$. 
In this paper, we consider a general $\cF$, which can be a class of parametric functions, nonparametric functions, regression trees, neural networks, etc.

We make some remarks on the above two settings from a pure modeling perspective. First, the agnostic setting does not require realizability and is more general than the realizable setting. Indeed, given any function class $\cF$, one can construct an \textit{induced} policy class $\Pi_{\cF}=\{\pi_f\mid f\in\cF\}$, thus any realizable contextual bandit problem can be reduced to an agnostic contextual bandit problem. Second, the realizable setting has its own merit, as the additional realizability assumption enables stronger performance guarantees: once the realizability assumption holds, the learner's competing policy $\pi_{f^*}$ is guaranteed to be \emph{globally optimal} (i.e., no policy can be better than $\pi_{f^*}$), thus small regret necessarily means large total reward. By contrast, in the no-realizability agnostic setting, the ``optimal policy in $\Pi$'' is not necessarily an effective policy if there are significantly more effective polices outside of $\Pi$. More comparisons between the two settings regarding theoretical tractability, computational efficiency and practical implementability will be provided in \S\ref{sec:lit}.

\subsection{Related Work}\label{sec:lit}
Contextual bandits have been extensively studied for two decades; see Chapter 5 of \cite{lattimore2018bandit} and Chapter 8 of \cite{slivkins2019introduction} for detailed surveys. Here we mention some important and closely related work.

\subsubsection{Agnostic Approaches} 
Papers studying contextual bandits in the agnostic setting aim to design general-purpose and computationally-tractable algorithms that are provably efficient for any given policy class $\Pi$ while avoiding the computational complexity of enumerating over $\Pi$ (as the size of $\Pi$ is usually extremely large). The primary focus of prior literature is on the case of general finite $\Pi$, as this is the starting point for further studies of infinite (parametric or nonparametric) $\Pi$. For this case, the EXP4-family algorithms (\citealt{auer2002nonstochastic,mcmahan2009tighter,beygelzimer2011contextual}) achieve the optimal $O(\sqrt{KT\log|\Pi|})$ regret but requires $\Omega(|\Pi|)$ running time at each round, which makes the algorithms intractable for large $\Pi$. In order to circumvent the $\Omega(|\Pi|)$ running time barrier, researchers (e.g., \citealt{langford2008epoch,dudik2011efficient,agarwal2014taming}) restrict their attention to \textit{oracle-based} algorithms that access the policy space only through an \textit{offline optimization oracle} --- specifically, an offline cost-sensitive classification oracle that solves
\begin{equation}\label{eq:csc}
\arg\max_{\pi\in\Pi}\sum_{s=1}^t \tilde{r}_{s}(\pi(x_s))
\end{equation}
for any given sequence of context and reward vectors $(x_1,\tilde{r}_1),\cdots,(x_t,\tilde{r}_t)\in\cX\times\mathbb{R}_{+}^{\cA}$. An \textit{oracle-efficient} algorithm refers to an algorithm whose number of oracle calls is polynomial in $T$ over $T$ rounds.

The first provably optimal oracle-efficient algorithm is the \texttt{Randomized UCB} algorithm of \cite{dudik2011efficient}, which achieves the optimal regret with $\widetilde{O}(T^6)$ calls to the cost-sensitive classification oracle. A breakthrough is achieved by the \texttt{ILOVETOCONBANDITS} algorithm in the celebrated work of \cite{agarwal2014taming}, where the number of oracle calls is significantly reduced to $\widetilde{O}(\sqrt{KT/\log|\Pi|})$. The above results are fascinating in theory because they enable a  ``online-to-offline reduction'' from contextual bandits to cost-sensitive classification, which is highly non-trivial for online learning problems in general. However, the practicability of the above algorithms is heavily restricted due to their reliance on the cost-sensitive classification oracle (\ref{eq:csc}), as this task is computationally intractable even for simple policy classes (\citealt{klivans2009cryptographic,agrawal2016linear}) and typically involves solving NP-hard problems. As a result, the practical implementations of the above classification-oracle-based algorithms typically resort to heuristics (\citealt{agarwal2014taming,bietti2018contextual}). Moreover, the above algorithms are memory hungry: since they must feed \textit{augmented} versions of the dataset (rather than the original version of the dataset) into the oracle, they have to repeatedly create auxiliary data and store them in memory. Therefore, these approaches may not perform well in practice (\citealt{bietti2018contextual}),  and are generally impractical for large-scale real-world deployments (\citealt{foster2018practical,foster2020beyond}).%

\subsubsection{Realizibility-based Approaches} 
In contrast to the agnostic setting where research primarily focuses on designing general-purpose algorithms that work for any given $\Pi$, a majority of research in the realizable setting tends to design specialized algorithms that work well for a particular parametrized family of $\cF$. Two of the dominant strategies for the realizable setting are upper confidence bounds (e.g., \citealt{filippi2010parametric,abbasi2011improved,chu2011contextual,li2017provably,li2019nearly}) and Thompson sampling (e.g., \citealt{agrawal2013thompson,russo2018tutorial}). While these approaches have been practically successful in several scenarios (\citealt{li2010contextual}), their theoretical guarantees and computational tractability critically rely on their strong assumptions on $\cF$, which restrict their usage  in other scenarios (\citealt{bietti2018contextual}).

To our knowledge, \cite{agarwal2012contextual} is the first paper studying contextual bandits with a general finite $\cF$, under the minimal realizability assumption. They propose a elimination-based algorithm, called \texttt{Regressor Elimination}, that achieves the optimal $\widetilde{O}(\sqrt{KT\log|\cF|})$ regret. However, their algorithm is computational inefficient, as it  enumerates over the whole function class and requires $\Omega(|\cF|)$ computational cost at each round (note that the size of $\cF$ is typically extremely large). The computational issues of \cite{agarwal2012contextual} are addressed by \cite{foster2018practical}, who propose an oracle-efficient contextual bandit algorithm  \texttt{RegCB}, which always accesses the function class through a weighted least squares regression oracle that solves
\begin{equation}\label{eq:wlsr}
\arg\min_{f\in\cF}\sum_{s=1}^t w_s(f(x_s,a_s)-y_s)^2
\end{equation}
for any given input sequence $(w_1,x_1,a_1,y_1),\cdots,(w_t,x_t,a_t,y_t)\in\mathbb{R}_{+}\times\cX\times\cA\times\mathbb{R}$. As \cite{foster2018practical} mention, the above oracle can often be solved efficiently and is very common in machine learning practice --- it is far more reasonable than the cost-sensitive classification oracle (\ref{eq:csc}).  However, unlike \texttt{Regressor Elimination}, the \texttt{RegCB} algorithm is not minimax optimal --- its worst-case regret could be as large as $\widetilde{\Omega}(T)$. Whether the optimal $\widetilde{O}(\sqrt{KT\log|\cF|})$ regret is attainable for an offline-regression-oracle-based algorithm remains unknown in the literature.

More recently, \cite{foster2020beyond} propose an algorithm that achieves the optimal regret for contextual bandits using an \textit{online} regression oracle. Their algorithm, called \texttt{SquareCB}, builds on the \texttt{A}/\texttt{BW} algorithm of \cite{abe1999associative} (see also the journal version \citealt{abe2003reinforcement}) originally developed for linear contextual bandits ---  specifically, \texttt{SquareCB}  replaces the ``Widrow-Hofff predictor'' used in the \texttt{A}/\texttt{BW} algorithm by a general online regression predictor, then follows the same probabilistic action selection strategy  as the \texttt{A}/\texttt{BW} algorithm. \cite{foster2020beyond} show that by using this simple strategy, contextual bandits can be  reduced to {online regression} in a black-box manner. While the implication that contextual bandits are no harder than online regression is important and insightful, online regression with a general function class itself is a challenging problem. Note that an online regression oracle has to provide robust guarantees for an arbitrary data sequence generated by an adaptive adversary, 
which may cause implementation difficulties when the function class $\cF$ is complicated %
--- while there is a beautiful theory characterizing the 
minimax regret rate of online regression with general function classes (\citealt{rakhlin2014online}), to our knowledge computational efficient algorithms are only known for specific function classes. For example, consider the case of a general finite $\cF$,  the online algorithm given by \cite{rakhlin2014online} actually requires $\Omega(|\cF|)$ computational cost at each round. Therefore, beyond the existing results of \cite{foster2020beyond}, a more thorough ``online-to-offline reduction'' from contextual bandits to offline regression is highly desirable.

\subsection{Technical Challenges and Our Approach}\label{sec:challenge}
Before we proceed to present our results, we would like to illustrate the key technical hurdles of using offline regression oracles to achieve the optimal regret for contextual bandits. We will then briefly explain how our approach overcomes these technical hurdles.

As was pointed out before, three excellent papers \cite{agarwal2012contextual,foster2018practical,foster2020beyond} have made important progress towards solving contextual bandits via regression approaches. Understanding the gap between the existing results and our desired result is important for  understanding the key technical hurdles. Below we discuss three challenges.%

\textbf{Computational hurdle.} \cite{agarwal2012contextual} propose a provably optimal but computational \text{inefficient} algorithm for contextual bandits with a general finite $\cF$. At each round $t$, their  algorithm maintains a subset $\cF_t\subset\cF$ based on successive elimination and solves a complicated optimization problem over $\cF_t$. Here, the key difficulty of using an offline regression oracle is that one cannot reformulate the complicated optimization problem over $\cF_t$ to a simple optimization problem like least squares regression, as the objective function is far more complicated than a sum of squares. This is also why using a square loss regression  oracle  is more challenging than using the offline cost-sensitive classification oracle (\ref{eq:csc}) --- one can understand the latter as a 0-1 loss oracle.%

\textbf{Statistical hurdle associated with constructing  confidence bounds.} \cite{foster2018practical} propose a computationally efficient confidence-bounds-based  algorithm  using an offline weighted least squares oracle. However, their algorithm only has statistical guarantees under some strong distributional assumptions. An important reason is that confidence-bounds-based algorithms typically rely on the ability of constructing \emph{shrinking} confidence intervals on \textit{each context}. While this is possible for a simple $\cF$ like a linear class, it is impossible for a general $\cF$. Here, the difficulty originates from the fact that all the  statistical learning guarantees for offline regression with a general $\cF$ require one to take an expectation over contexts. In other words, effective per-context statistical guarantees are generally impossible for an offline regression oracle.

\textbf{Statistical hurdle associated with analyzing dependent actions.} \cite{foster2020beyond} propose an optimal and efficient contextual bandit algorithm assuming access to an online regression oracle, which is quite different from an offline regression oracle. Statistically, the difference between offline and online regression oracles is that, offline regression oracles only assume statistical guarantees for an i.i.d. data sequence (see \S\ref{sec:extension} for our definition of a general offline regression oracle), while  online regression oracles assume statistical guarantees for  an arbitrary data sequence possibly generated by an adaptive adversary. Evidently, access to an online regression oracle is a stronger assumption than access to an offline regression oracle. As \cite{foster2020beyond} mention, their algorithm requires an online regression oracle because ``the analysis critically uses that the regret bound (of the online regression oracle) holds when the actions $a_1,\dots,a_T$ are chosen adaptively, since actions selected in early rounds are used by \texttt{SquareCB} to determine the action distribution at later rounds.'' That is, the technical hurdle of using an offline regression oracle here is that the algorithm's action sequence is not i.i.d. --- since offline regression oracles are designed for i.i.d. data, it is unclear  how one can deal with dependent actions when one only has access to an offline regression oracle. We note that this hurdle lies at the heart of the ``exploration-exploitation trade-off'' --- essentially, any efficient algorithm's actions must be highly dependent, as they are simultaneously used for exploration and exploitation.

\subsubsection{Our Resolution}

We address the three technical hurdles in \S\ref{sec:challenge} in a surprisingly elegant way. Specifically, we derive an algorithm that accesses the offline regression oracle in a mostly ``naive'' way, without constructing any explicit optimization problems or confidence bounds, thus gets around the first two hurdles simultaneously; further, we overcome the third hurdle by establishing a framework to analyze our algorithm and prove its statistical optimality  --- in particular, we face the complex dynamics of evolving dependent actions, but analyze them through a different lens (the ``dual interpretation'' in \S\ref{sec:analysis}), and establish a series of sufficient conditions for optimal contextual bandit learning under this lens. The final algorithm is simple, but the ideas behind it are quite deep and are supported by novel analysis. The algorithmic details will be presented in \S\ref{sec:alg} and \S\ref{sec:extension} and the key ideas will be explained in \S\ref{sec:analysis}.

Our approach builds on (and reveals connections between) two lines of research in the contextual bandit literature: (i) a celebrated theory of optimal contextual bandit learning in the  agnostic setting using a (seemingly unavoidable) classification oracle, represented by \cite{dudik2011efficient} (the ``monster paper'') and \cite{agarwal2014taming} (``taming the monster''); (ii) a simple probabilistic selection strategy mapping the predicted rewards of actions to the probabilities of actions, pioneered by \cite{abe1999associative} (see also \citealt{abe2003reinforcement}) and extended by \cite{foster2020beyond}. %
In particular, we rethink the philosophy behind \cite{dudik2011efficient} and \cite{agarwal2014taming}, reform it with our own understanding of the value of realizability, and come up with a new idea of ``bypassing'' the classification oracle under realizability --- our algorithm is essentially a  consequence of this new idea; see \S\ref{sec:born}. Interestingly, our derived algorithm 
turns out to use essentially the same probabilistic selection strategy as \cite{abe1999associative} and \cite{foster2020beyond} --- this is surprising, as  the idea behind the derivation of our algorithm is very different from the ideas behind \cite{abe1999associative} and \cite{foster2020beyond}. This suggests that this simple probabilistic selection strategy might be more intriguing and more essential for bandits than previously understood, and we believe that it is worth further attention from the bandit community. We hope that our work, together with \cite{abe1999associative} and \cite{foster2020beyond}, can provide diverse perspectives on how to understand  this strategy. 

As a final remark, we emphasize that compared with each line of research that we mention above, our approach has new contributions beyond them which seem necessary for our arguments to hold. We will elaborate on such new contributions in the rest of our article.%

\subsection{Organization and Notations}\label{sec:notation}
The rest of the paper is organized as follows. For pedagogical reasons, we first present our results in the case of a general finite $\cF$ in \S\ref{sec:alg}, where  we  introduce our algorithm and state its  theoretical guarantees.  %
In \S\ref{sec:extension}, we extend our results to the general setting and discuss several important consequences. In \S\ref{sec:analysis}, we present our regret analysis and explain the ideas behind our algorithm.
We conclude our paper in \S\ref{sec:conclusion}. All the proofs of our results are deferred to the appendix.

Throughout the paper, we use $O(\cdot)$ to hide constant factors, and $\widetilde{O}(\cdot)$ to hide $\mathrm{polylog}(T)$ factors. Given $\mcD$, let $\cD$ denote the marginal distribution over $\cX$. We use $\sigma(Y)$ to denote the $\sigma$-algebra generated by a random variable $Y$, and use $\cB(E)$ to denote the Borel $\sigma$-algebra  on a set $E$. An \emph{action selection kernel} $p:\cB(\cA)\times\cX\rightarrow[0,1]$ is defined as a probability kernel such that $p(a\mid x)$ specifies the probability of selecting action $a\in\cA$ given context $x\in\cX$; let $\cP$ be the space of all action selection kernels. We use $\N$ to denote the set of all positive integers, and $\mathbb{R}_{+}$ to denote the set of all non-negative real numbers. Without loss of generality, we assume that $|\cF|\ge4$.

\section{Algorithm and Guarantees}\label{sec:alg}
Following previous work (\citealt{dudik2011efficient,agarwal2012contextual,agarwal2014taming}), we start with the case of a general finite $\cF$, as this is the  starting point for further studies of an infinite $\cF$. For this case, the ``gold standard'' is an algorithm that achieves $\widetilde{O}(\sqrt{KT\log|\cF|})$ regret with the total number of oracle calls being polynomial/sublinear in $T$ (see \citealt{agarwal2012contextual,foster2018practical}). As for the oracle, we assume access to the following  \emph{least squares regression oracle} that solves
\begin{equation}\label{eq:lsr}
\arg\min_{f\in\cF}\sum_{s=1}^t (f(x_s,a_s)-y_s)^2
\end{equation}
for any input sequence $(x_1,a_1,y_1),\cdots,(x_t,a_t,y_t)\in\cX\times\cA\times[0,1]$. Without loss of generality\footnote{If the oracle is allowed to return a random solution (when there are multiple optimal solutions), then we can simply incorporate such randomness into the history when we define $\Upsilon_t$ in Appendix \ref{app:def}, and all our proofs will still hold.}, we assume that the oracle (\ref{eq:lsr}) always returns the same solution for two identical input sequences. Note that the above least squares  oracle (\ref{eq:lsr}) is a concrete optimization oracle and is simpler than the weighted one (\ref{eq:wlsr}) assumed in \cite{foster2018practical}, as it does not need to consider the weights.

We remark that our reduction is not restricted to this setup --- in \S\ref{sec:extension}, we will extend all our results to the general setting where both  $\cF$ and the offline regression oracle are generic. Still, the above setup is good for illustrating our results, and allows direct comparisons to the ``gold standard.''

\subsection{The Algorithm}

We present our algorithm, ``FAst Least-squares-regression-oracle CONtextual bandits'' (\texttt{FALCON}), in Algorithm \ref{alg:cbor} (a generalized version of this algorithm will be provided in \S\ref{sec:extension}). The algorithm is very simple and follows the same general template as the \texttt{A}/\texttt{BW} algorithm of \cite{abe1999associative} and the \texttt{SquareCB} algorithm of \cite{foster2020beyond}, with the main difference lying in using a different oracle to generate predictions. We also add a few useful ingredients, including an epoch schedule and a changing learning rate. See the description of the algorithm below. %

\begin{algorithm}[htbp]
\caption{FAst Least-squares-regression-oracle CONtextual bandits (\texttt{FALCON})}
\label{alg:cbor}
{\textbf{input} epoch schedule $0=\tau_0<\tau_1<\tau_2<\cdots$, confidence parameter $\delta$, tuning parameter $c$}\\
\vspace*{-2em}
\begin{algorithmic}[1]
\FOR{epoch $m=1,2,\dots$}
    \STATE{Let $\gamma_m=c\sqrt{K\tau_{m-1}/\log(|\cF|\log(\tau_{m-1})m/\delta)}$ (for epoch 1, $\gamma_1=1$).}
    \STATE{Compute $\widehat{f}_m=\arg\min_{f\in\cF}\sum_{t=1}^{\tau_{m-1}}(f(x_t,a_t)-r_t(a_t))^2$ via the \textbf{offline least squares oracle}.}
    \FOR{round $t=\tau_{m-1}+1,\cdots,\tau_m$}
        \STATE{Observe context $x_t\in\cX$}.
        \STATE{Compute $\widehat{f}_m(x_t,a)$ for each action $a\in\cA$. Let $\widehat{a}_t=\max_{a\in\cA}\widehat{f}_m(x_t,a)$}. Define
        \begin{equation*}
        p_t(a)=\begin{cases}
        \frac{1}{K+\gamma_m\left(\widehat{f}_m(x_t,\widehat{a}_t)-\widehat{f}_m(x_t,a)\right)},&\text{for all }a\ne \widehat{a}_t,\\
        1-\sum_{a\ne\widehat{a}_t}p_t(a),&\text{for }a=\widehat{a}_t.
        \end{cases}
        \end{equation*}\vspace*{-0.2in}
        \STATE{Sample $a_t\sim p_t(\cdot)$ and observe reward $r_t(a_t)$.}
    \ENDFOR
\ENDFOR
\end{algorithmic}
\end{algorithm}

Our algorithm runs in an epoch schedule to reduce oracle calls, i.e., it only calls the oracle at certain pre-specified rounds $\tau_1,\tau_2,\tau_3,\dots$. For $m\in\N$, we refer to the rounds from $\tau_{m-1}+1$ to $\tau_m$ as epoch $m$. %
As a concrete example, consider $\tau_m=2^m$, then for any (possibly unknown) $T$, our algorithm runs in $O(\log T)$ epochs. As another example, when $T$ is known, consider $\tau_m=\left\lfloor 2T^{1-2^{-m}}\right\rfloor$, then our algorithm runs in $O(\log\log T)$ epochs. We allow very general epoch schedules; in particular, calling the oracle more frequently does not affect the regret analysis.

At the start of each epoch $m$, our algorithm makes two updates. First, it updates a (epoch-varying) learning rate $\gamma_m\simeq \sqrt{K\tau_{m-1}/\log(|\cF|/\delta)}$, which aims to strike a balance between exploration and exploitation. Second, it computes a ``greedy'' predictor $\widehat{f}_m$ from $\cF$ that minimizes the empirical square loss $\sum_{t=1}^{\tau_{m-1}}(f(x_t,a_t)-r_t(a_t))^2$. This predictor can be computed via a single call to the offline least squares regression oracle --- notably, $\min_{f\in\cF}\sum_{t=1}^{\tau_{m-1}}(f(x_t,a_t)-r_t(a_t))^2$ is almost the best way that we can imagine for our oracle to be called, with no augmented data generated, no weights maintained, and no additional optimization problem constructed.

The decision rule in epoch $m$ is then completely determined by $\gamma_m$ and $\widehat{f}_m$. For each round $t$ in epoch $m$, given a context $x_t$, the algorithm uses $\widehat{f}_m$ to predict each action's reward and finds a greedy action $\widehat{a}_t$ that maximizes the predicted reward. Yet the algorithm does not directly select $\widehat{a}_t$ --- instead, it randomizes over all actions according to a probabilistic selection strategy that picks each action other than $\widehat{a}_t$ with probability roughly inversely proportional to how much worse it is predicted to be as compared with  $\widehat{a}_t$, as well as roughly inversely proportional to the learning rate $\gamma_m$. The effects of this strategy are twofold. First, \textit{at each round}, by assigning the greedy action the highest probability and each non-greedy action a probability roughly inverse to the predicted reward gap, we ensure that the better an action is predicted to be, the more likely it will be selected. Second, \textit{across different epochs}, by controlling the probabilites of non-greedy actions roughly inverse to the gradually increasing learning rate $\gamma_m$, we ensure that the algorithm ``explores more'' in the beginning rounds where the learning rate is small, and gradually ``exploits more'' in later rounds where the learning rate becomes larger --- this is why we view our learning rate as a sequential balancer between exploration and exploitation.

\textbf{Algorithmic components and comparisons with literature.}  
\texttt{FALCON} is a very simple algorithm, and can be viewed as a  combination of three algorithmic components: (i) an epoch schedule, (ii) the greedy use of an offline least squares regression oracle, and (iii) a probabilistic selection strategy that maps reward predictions to action probabilities, controlled by an epoch-varying learning rate. While each component alone is not new in the literature, %
the combination of the above three components has not been considered in the literature, and it is far from obvious that this particular combination  should be effective. In fact, it is quite surprising  that such a simple algorithm  would work  well for general contextual bandits. While there  is definitely more to this  algorithm than meets the eye (we will explain the essential idea behind \texttt{FALCON} in \S\ref{sec:bypass} and \S\ref{sec:born}), let us first give a few quick comments on component (ii) and (iii), and compare them to existing literature.

We start from component (iii). As we mention before, the idea of mapping the predicted action rewards to action probabilities via an ``inverse proportional to the gap'' rule is not new: such a probabilistic selection strategy is firstly proposed  by \cite{abe1999associative}  in their study of linear contextual bandits, and recently adopted by \cite{foster2020beyond} in their reduction from contextual bandits to \textit{online} regression. Compared with the existing strategy used in \cite{abe1999associative} and \cite{foster2020beyond},
the strategy that we use here has a notable difference: %
while the above two papers adopt a constant learning rate $\gamma$ that does not change in the running process of their algorithms%
, we appeal to an epoch-varying (or time-varying) learning rate $\gamma_m\simeq\sqrt{K\tau_{m-1}/\log(|\cF|/\delta)}$ that gradually increases as our algorithm proceeds.
This epoch-varying learning rate plays a important role in our statistical analysis, as the proof of our regret guarantee  relies on an inductive argument which requires the learning rate to change carefully with respect to epochs and gradually increase over time; see \S\ref{sec:ocbl}.

\emph{Remark. }{While such an epoch-varying learning rate is not necessary when $T$ is known in advance and the oracle calls are ``frequent'' enough, an epoch-varying learning rate brings certain benefits to the algorithm: first, in the case of unknown $T$, it is required; second, in the case of known $T$, it is necessary whenever one seeks to control the total number of oracle calls within $o(\log T)$ (a fixed learning rate could lead to sub-optimal regret in this setting); third, in our analysis it always leads to tighter regret bounds with better dependence on logarithmic factors. As a result, it seems that an epoch-varying learning rate always dominates a fixed learning rate in our problem.}

Component (ii) of our algorithm is particularly interesting. %
Indeed, our algorithm makes predictions in a surprisingly simple and straightforward way: it always picks the greedy predictor and
directly applies it on contexts without any modification --- that is, in terms of making predictions, the algorithm is fully greedy. This seems to contradict the conventional idea that greedy-prediction-based algorithms are typically sub-optimal (e.g., \citealt{langford2008epoch}), and is in sharp contrast to previous elimination-based algorithms (e.g., \citealt{dudik2011efficient,agarwal2012contextual}) and confidence-bounds-based algorithms (e.g., \citealt{abbasi2011improved,chu2011contextual}) ubiquitous in the bandit literature, which spend a lot of efforts and computation resources maintaining complex  confidence intervals, version spaces, or distributions over predictors. Even when one thinks about the algorithms of \cite{abe1999associative} and \cite{foster2020beyond} which are similar to ours, one can find that they appeal to more robust predictors: \cite{abe1999associative} appeal to the ``Widrow-Hofff predictor'' (equivalent to an online gradient descent oracle) and  \cite{foster2020beyond} appeal to a general online regression oracle. Both of their analysis critically relies on the \textit{online} nature of their oracles, i.e., the oracles can  efficiently minimize regret  against an adaptive adversary --- essentially, this means that a portion of the heavy lifting regarding the exploration-exploitation trade-off is taken care of by the online oracles, not the algorithms. While seemingly counter-intuitive, we claim that making ``naive'' greedy predictions is sufficient for optimal contextual bandit learning, which means that our oracle does not care about the exploration-exploitation trade-off at all. This surprising finding suggests that a rigorous analysis of our algorithm should contain some new ideas  beyond existing bandit literature. %
Indeed, we will provide a quite interesting analysis of our algorithm in \S\ref{sec:analysis}, which seem to be conceptually novel.

\emph{Remark. }Readers who are interested in the difference between an offline oracle and an online oracle may compare the regret analysis approach in this paper with the approaches in \cite{abe1999associative} and \cite{foster2020beyond}. %
The analysis of \cite{abe1999associative} and \cite{foster2020beyond} is essentially per-round analysis: at each round, the instantaneous bandit regret is
upper bounded by the instantaneous online regression regret, with no structure shared across different rounds, so the final regret bound  follows from taking a sum over all rounds. By contrast, our analysis has to deal with the shared structure across different rounds, i.e., we have to figure out how the exploration that occurred in early rounds benefits the exploitation in later rounds.

\subsection{Theoretical Guarantees}
We show that the simple algorithm \texttt{FALCON} enjoys strong performance guarantees.

\textbf{Statistical optimality.}  Define $m(T):=\min\{m\in\N:T\le\tau_m\}$, which is the total number of epochs that Algorithm \ref{alg:cbor} executes. 
The regret guarantee of Algorithm \ref{alg:cbor} is stated in Theorem \ref{thm:finite}.  The proof is deferred to Appendix \ref{app:thm1}. We will elaborate on the key ideas of the analysis in \S\ref{sec:analysis}. %
\begin{theorem}\label{thm:finite} Consider an epoch schedule such that $\tau_{m}\le2\tau_{m-1},\,\forall m>1$ and $\tau_1\le2$. Let $c=1/30$.
For any $T\in\N$, with probability at least $1-\delta$, the regret of Algorithm \ref{alg:cbor} after $T$ rounds is at most
$$
O\left(\sqrt{KT\log(|\cF|m(T)/\delta)}\right).
$$
\end{theorem}
When $\tau_m=2^m$, the above upper bound is $O\left(\sqrt{KT\log(|\cF|\log T/\delta)}\right)$, which removes a  superfluous $\sqrt{\log T}$ factor in the regret upper bound of \cite{agarwal2012contextual} (attained by an \emph{inefficient} algorithm), and matches the lower bound proved by  \cite{agarwal2012contextual} up to a constant or $\sqrt{\log\log T}$ factor. The \texttt{FALCON} algorithm is thus statistically optimal. %

\textbf{Computational efficiency.} 
Consider the epoch schedule $\tau_m=2^m$, $\forall m\in\N$. For any possibly unknown $T$, our algorithm runs in $O(\log T)$ epochs, and in each epoch our algorithm only calls the oracle once. Therefore, our algorithm's computational complexity is $O(\log T)$ calls to a least squares regression oracle  across all $T$ rounds (plus $O(K)$ additional cost per round). This leads to potential advantages over existing algorithms. Note that \texttt{ILOVETOCONBANDITS} requires $\widetilde{O}(\sqrt{KT/\log(|\cF|/\delta)})$ calls to an offline cost-sensitive classification oracle, and \texttt{SquareCB} requires $O(T)$ calls to an {online} regression oracle --- compared with our algorithm, both of them require considerably more calls to  harder-to-implement oracles (as far as a general finite $\cF$ is concerned). Also, since a general finite $\cF$ is not a convex function class, \texttt{RegCB} requires $O(T^{3/2})$ calls to a weighted least squares regression oracle for this setting --- this is also much slower than our  algorithm.

When the total number of rounds $T$ is known to the learner, we can make the computational cost of \texttt{FALCON} even lower. For any $T\in\N$, consider an epoch schedule $\tau_m=\left\lfloor{2T^{1-2^{-m}}}\right\rfloor$, $\forall m\in\N$ (similar to \citealt{cesa2014regret}). Then \texttt{FALCON} will run in $O(\log\log T)$ epochs, calling the oracle for only $O(\log\log T)$ times over $T$ rounds. In this case, we still have the same regret guarantee (up to a $\log\log T$ factor); see Corollary \ref{cor:finite} below. The proof  can be found in Appendix \ref{app:final}.

\begin{corollary}\label{cor:finite}
For any $T\in\N$, consider an epoch schedule $\tau_m=\left\lfloor  2T^{1-2^{-m}}\right\rfloor$, $\forall m\in\N$ and let $c=1/30$. With probability at least $1-\delta$, the regret of Algorithm \ref{alg:cbor} after $T$ rounds is at most
$$
O\left(\sqrt{KT\log(|\cF|\log T/\delta)}\log\log T\right).
$$
\end{corollary}

\section{General Offline Regression Oracles}\label{sec:extension}

We now extend our results to the general setting where $\cF$ is generic (possibly infinite). While  we can still assume a least squares regression oracle as before (which corresponds to the \textit{empirical risk minimization} (ERM) procedure under square loss in offline supervised learning), for different $\cF$, some other types of offline regression procedures (e.g., regularized least squares like Ridge and Lasso, or logistic regression) may be preferred. Moreover, even for a function class where least squares regression is preferred, one may not want to solve the square loss minimization problem exactly, and an oracle that allows optimization error may be preferred. Therefore, in this section, we state our results in a more general way: we assume access to an arbitrary offline regression oracle with a generic statistical learning guarantee, and design an algorithm that makes calls to this arbitrary  oracle and utilizes its statistical learning guarantees. Recall that the goal of this paper is to accomplish an online-to-offline reduction from contextual bandits to offline regression. So ultimately, we want to provide a universal and optimal ``offline-to-online converter,'' such that existing machinery of supervised learning with general function classes can be automatically translated into contextual bandits with general function classes. 

In what follows, we introduce the notion of a \emph{general offline regression oracle}.

Given a general function class $\cF$, a general {offline regression oracle} associated with $\cF$, denoted by $\texttt{OffReg}_{\cF}$, is defined as a procedure that generates a predictor $\widehat{f}:\cX\times\cA\rightarrow\mathbb{R}$ based on input data\footnote{Without loss of generality, assume that $\texttt{OffReg}_{\cF}$ always returns the same predictor for two identical input sequences.} and $\cF$ (note that $\widehat{f}$ need not be in $\cF$). In statistical learning theory, the quality of $\widehat{f}$ is typically measured by its ``out-of-sample error,'' i.e., its expected error on \emph{random} and \emph{unseen} test data. We make the following generic assumption on the statistical learning guarantee of $\texttt{OffReg}_{\cF}$.
\begin{assumption}\label{assump:oracle}
Let $p$ be an arbitrary action selection kernel (see \S\ref{sec:notation} for the definition). 
Given $n$ training samples of the form $(x_i,a_i;r_i(a_i))$ {independently and identically drawn} according to  $(x_i,r_i)\sim\mcD$, $a_i\sim p(\cdot\mid x_i)$, the offline regression oracle $\texttt{OffReg}_{\cF}$ returns a predictor $\widehat{f}:\cX\times\cA\rightarrow\mathbb{R}$. For any $\delta>0$, with probability at least $1-\delta$, we have
\[
\E_{x\sim\cD,a\sim p(\cdot\mid x)}\left[(\widehat{f}(x,a)-f^*(x,a))^2\right]\le\cE_{\cF,\delta}(n).
\]
\end{assumption}
The offline learning guarantee $\cE_{\cF,\delta}(n)$ is a function that decreases with $n$, which bounds the squared {$L_2$ distance} between $\widehat{f}$ and $f^*$ on the test data (generated from the same distribution as the training data). Under realizability (i.e., $f^*\in\cF$), this squared distance corresponds to the \emph{estimation error} or \emph{excess risk}  of $\widehat{f}$ (under square loss, or more broadly, strongly convex loss\footnote{The {estimation error} / {excess risk} is defined as $\E[\ell(\widehat{f}(x,a),r(a))]-\inf_{f\in\cF}\E[\ell(f(x,a),r(a))]$ for a general loss function $\ell$. When $\ell$ is the {square loss}, it equals to $\E[(\widehat{f}(x,a)-f^*(x,a))^2]$ under realizability. Moreover, any excess risk bound under a strongly convex loss such as the {log loss} implies an upper bound on  $\E[(\widehat{f}(x,a)-f^*(x,a))^2]$ under realizability.}). 
Note that characterizing sharp estimation error / excess risk bounds and designing efficient algorithms to attain such bounds are among the most central tasks in statistical learning.%

The above notion of the offline regression oracle, though being very natural, appears to be new in the contextual bandit literature. In particular, it is not restricted to the least squares oracle (thus finds broader applications), and it is strictly easier to implement than the online regression oracle of \cite{foster2020beyond} which has to deal with sequential data generated by an adaptive adversary. Indeed, any oracle satisfying the requirement of \cite{foster2020beyond} can be easily converted to an oracle that satisfy our  Assumption \ref{assump:oracle}.

Reducing contextual bandits to the above general offline regression oracle brings many important advantages, which will be discussed after our reduction is presented; see \S\ref{sec:advantages}. %

\subsection{Algorithm and Guarantees}

We provide an algorithm, called \texttt{FALCON+}, in Algorithm \ref{alg:falcon+}. The key differences between Algorithm \ref{alg:falcon+} and Algorithm \ref{alg:cbor} lie in step 2 and step 3. In step 2, we define a new epoch-varying learning rate based on the offline learning guarantee of  $\texttt{OffReg}_{\cF}$ --- this is a direct generalization of the learning rate defined in Algorithm \ref{alg:cbor}. In step 3, instead of feeding all the previous data into the oracle, we {only feed the data in epoch $m-1$ into the oracle}. We make two comments here. First, while we do not feed all the previous data into the oracle any more, this is still a greedy-type call to the offline oracle, as we do not make any exploration consideration in this step. Second, the strategy of only feeding the data in the last epoch into the oracle is purely due to technical reasons (i.e., Assumption \ref{assump:oracle} requires i.i.d. data), as we want to avoid a more complicated discussion of  martingales. Note that as a consequence of this strategy, our algorithm must run in gradually increasing epochs, e.g., $\tau_m=2^m$ or $\tau_m=\left\lfloor2 T^{1-2^{-m}}\right\rfloor$.

\begin{algorithm}[htbp]
\caption{FAst generaL-offline-regression-oracle CONtextual bandits (\texttt{FALCON+})}
\label{alg:falcon+}
{\textbf{input} epoch schedule $0=\tau_0<\tau_1<\tau_2<\cdots$, confidence parameter $\delta$, tuning parameter $c$}\\
\vspace*{-2em}
\begin{algorithmic}[1]
\FOR{epoch $m=1,2,\dots$}
    \STATE{Let $\gamma_m=c\sqrt{K/\cE_{\cF,\delta/(2m^2)}(\tau_{m-1}-\tau_{m-2})}$ (for epoch 1, $\gamma_1=1$).}
    \STATE{Feed (\textbf{only}) the data in epoch $m-1$, i.e., \[(x_{\tau_{m-2}+1},a_{\tau_{m-2}+1};r_{\tau_{m-2}+1}(a_{{\tau_{m-2}+1}})),\cdots, (x_{\tau_{m-1}},a_{\tau_{m-1}};r_{\tau_{m-1}}(a_{{\tau_{m-1}}}))\] into  the \textbf{offline regression oracle} $\texttt{OffReg}_{\cF}$ and obtain $\widehat{f}_m$ (for epoch 1, $\widehat{f}_1\equiv0$).}
    \FOR{round $t=\tau_{m-1}+1,\cdots,\tau_m$}
        \STATE{Observe context $x_t\in\cX$}.
        \STATE{Compute $\widehat{f}_m(x_t,a)$ for each action $a\in\cA$. Let $\widehat{a}_t=\max_{a\in\cA}\widehat{f}_m(x_t,a)$}. Define
        \begin{equation*}
        p_t(a)=\begin{cases}
        \frac{1}{K+\gamma_m\left(\widehat{f}_m(x_t,\widehat{a}_t)-\widehat{f}_m(x_t,a)\right)},&\text{for all }a\ne \widehat{a}_t,\\
        1-\sum_{a\ne\widehat{a}_t}p_t(a),&\text{for }a=\widehat{a}_t.
        \end{cases}
        \end{equation*}\vspace*{-0.2in}
        \STATE{Sample $a_t\sim p_t(\cdot)$ and observe reward $r_t(a_t)$.}
    \ENDFOR
\ENDFOR
\end{algorithmic}
\end{algorithm}

Recall that $m(T)$ is the total number of epochs that Algorithm \ref{alg:falcon+} executes. The regret guarantee of Algorithm \ref{alg:falcon+} is stated in Theorem \ref{thm:inf}. The proof of Theorem 2 is deferred to Appendix \ref{app:thm2}.

\begin{theorem}\label{thm:inf}
Consider an epoch schedule such that $\tau_m\ge2^m$ for $m\le m(T)$ and let $c=1/2$. Without loss of generality, assume that $\gamma_1\le\cdots\le\gamma_{m(T)}$. For any $T\in\N$, with probability at least $1-\delta$, the regret of Algorithm \ref{alg:falcon+} after $T$ rounds is at most
\begin{equation}\label{eq:bound}
O\left(\sqrt{K}\sum_{m=2}^{m(T)}\sqrt{\cE_{\cF,\delta/(2m^2)}(\tau_{m-1}-\tau_{m-2})}(\tau_m-\tau_{m-1})\right).
\end{equation}
\end{theorem}

The above regret bound is general and it typically has the same rate as $O\left(\sqrt{K{\cE_{\cF,\delta/\log T}(T)}}T\right)$. Therefore, given an arbitrary offline regression oracle with an arbitrary estimation error guarantee $\cE_{\cF,\delta}(\cdot)$, we know that our algorithm's regret is upper bounded by $O\left(\sqrt{K{\cE_{\cF,\delta/\log T}(T)}}T\right)$. 

\begin{example}[Statistical Optimality of \texttt{FALCON+}]
Consider a general, potentially nonparametric function class $\cF$ whose \textit{empirical entropy} is $O(\varepsilon^{-p})$, $\forall \varepsilon>0$ for some constant $p>0$.  \cite{yang1999information} and \cite{rakhlin2017empirical} provide several offline regression oracles that achieve the optimal $\cE_{\cF}(n)=O(n^{-2/(2+p)})$ estimation error rate. By letting $\tau_{m}=2^m$ for $m\in\N$, the regret of \texttt{FALCON+} is upper bounded by $O(T^{\frac{1+p}{2+p}}\log T)$ when one ignores the dependence on $K$. Combined with an $\widetilde{\Omega}(T^{\frac{1+p}{2+p}})$ lower bound proved in \cite{foster2020beyond}, we know that \texttt{FALCON+} is rate-optimal as long as the offline regression oracle is rate-optimal. We thus accomplish a universal and optimal reduction from contextual bandits to offline regression. %
We note that the above result also helps characterize the minimax regret rate of stochastic contextual bandits with a general, potentially nonparametric $\cF$. Note that \cite{foster2020beyond} have already provided such a characterization, under a \emph{tensorization} assumption (see their Section 3 for details). We remove this assumption, as the $O(T^{\frac{1+p}{2+p}}\log T)$ upper bound implied by our Theorem \ref{thm:inf} recovers Theorem 3 of \cite{foster2020beyond}, without assuming tensorization. %
\end{example}

\begin{example}[Linear Contextual Bandits]Consider the linear contextual bandit setting of \cite{chu2011contextual} with stochastic contexts. This corresponds to setting $\cF$ to be the linear class $$\cF=\{(x,a)\mapsto \theta^\top x_a \mid \theta\in\mathbb{R}^d, \|\theta\|_2\le 1\},$$
where $x=(x_a)_{a\in\cA}$, $x_a\in\mathbb{R}^d$ and $\|x_a\|_2\le1$.
In this case, by using the least squares regression oracle, \texttt{FALCON+} achieves the regret $O(\sqrt{KT(d+\log T)})$. Compared with the best known upper bound for this problem, $\mathrm{poly}(\log\log KT)O(\sqrt{Td\log T\log K})$ in \cite{li2019nearly}, the regret bound of \texttt{FALCON+} has worse dependence on $K$ (which seems to come from the employed sampling strategy), but saves a $\sqrt{\log T}$ factor, which means that \texttt{FALCON+} improves the best known regret upper bound for this problem when $K<<T$. To the best of our knowledge, this is the first time that an algorithm gets over the $\Omega(\sqrt{Td\log T})$ barrier for this problem --- notably, our new upper bound even ``breaks'' the $\Omega(\sqrt{T d\log T\log K})$ lower bound proved in \cite{li2019nearly}. The caveat here is that \cite{li2019nearly} study the setting where contexts are chosen by an \textit{oblivious} adversary, while we are considering the setting where contexts are stochastic. Our finding that the $\Omega(\sqrt{Td\log T})$ barrier does not exist for linear contextual bandits with stochastic contexts is quite interesting. %
\label{ex:linear}
\end{example}

\begin{example}[Contextual Bandits with Neural Networks] Deriving provable performance guarantees for neural networks is an active area of research. Here we use a recent result of \cite{farrell2021deep} to illustrate how estimation error bounds for deep neural networks can be  translated into contextual bandits. Specifically, let $\cF=\mathcal{G}^K$, $\mathcal{G}$ be the class of \emph{Multi-Layer Perceptrons} (MLP) as described in Section 2.1 of \cite{farrell2021deep}, and $f^*(x,a)=g_a^*(x)$ for $x\in\cX, a\in\cA$. Assume that $\cD$ is a continuous distribution on $[-1,1]^d$ and  $g_1^*,\dots,g_K^*$ lie in a Sobolev ball with smoothness $\beta\in\N$. By Theorem 1 of \cite{farrell2021deep}, the deep MLP-ReLU network estimator attains $\widetilde{O}(n^{-\frac{\beta}{\beta+d}})$ estimation error. Consequently, \texttt{FALCON+} attains $\widetilde{O}(T^\frac{\beta+2d}{2\beta+2d})$ regret by using this estimator as the offline regression oracle (we omit the dependence on $K$ here). The above result is new, but cannot be directly compared with existing results on ``neural contextual bandits'' (e.g., \citealt{zhou2020neural}), as the model assumptions  are very different.%
\end{example}

In general, one can set $\cF$ to be any parametric or nonparametric function class, e.g., high-dimensional parametric class, Lipschitz function class, reproducing kernel Hilbert space, and regression-tree-based or random-forest-based class. For any function class $\cF$, we can obtain a practical algorithm achieving the optimal regret for the corresponding contextual bandit problem,  as long as we can find a computationally-efficient and statisitcally-optimal offline regression oracle. This usually leads to faster algorithms with improved regret bounds. In particular, our regret upper bounds' dependence on $T$ is usually better than previous upper bounds in the literature, thanks to the fact that  we lose very little in terms of dependence on $T$ when we directly convert an offline estimation error bound to a regret bound. Moreover, our results enable people to  tackle broader classes of new contextual bandit problems, such as contextual bandits with heavy-tailed rewards, which will be discussed shortly. %

\subsection{Discussion}\label{sec:advantages}
We discuss some interesting observations regarding our Assumption \ref{assump:oracle} and Theorem \ref{thm:inf}, which further demonstrate the generality of our results.%

\textbf{Exact solutions to ERM are not required.} An important advantage of Assumption \ref{assump:oracle} is that it does not pose any restriction on how the predictor $\widehat{f}$ is generated, thus does not require one to use ERM or exactly solve ERM. This implies that the offline predictor $\widehat{f}$ can be obtained by running iterative optimization algorithms like (stochastic) gradient descent, and its computation can be implemented in an online/streaming fashion on large datasets, which is an important
consideration in modern machine learning practice. In other words, $\widehat{f}$ can be computed via various methods, %
and the optimization error of $\widehat{f}$ is already included in the offline learning guarantee $\cE_{\cF,\delta}(n)$.

\textbf{Exact realizability is not required.} Another observation is that some approximation error can also be included in $\cE_{\cF,\delta}(n)$, which enables one to consider some  relaxed notions of realizability. Note that the proof of Theorem \ref{thm:inf} does not rely on the realizability assumption --- the proof only relies on Assumption \ref{assump:oracle}, which is well-defined even if $f^*\notin\cF$. This means that Algorithm \ref{alg:falcon+} and
the regret bound (\ref{eq:bound}) do not really require $f^*\in\cF$ --- all they need is a \emph{known} guarantee $\cE_{\cF,\delta}(n)$ which correctly upper bounds the population $L_2$ distance between $\widehat{f}$ and $f^*$.  
As a consequence, our results readily extend to the setting where realizability only holds approximately  up to a \emph{known} misspecification error $\epsilon$ (\citealt{van2019comments,lattimore2020learning,foster2020beyond}). Specifically, suppose that $f^*\notin\cF$ but there exists a function $\tilde{f}\in\cF$ that is close to $f^*$ in the sense that $\sup_{x,a}|{\tilde{f}(x,a)-f^*(x,a)}|\le\epsilon$, then we can deduce that
\[
\E[(\widehat{f}(x,a)-f^*(x,a))^2]\le\epsilon^2+\underbrace{\E[(\widehat{f}(x,a)-r(a))^2]-\inf_{f\in\cF}\E[(f(x,a)-r(a))^2]}_{\text{estimation error}}
\]
This means that one can take $\cE_{\cF,\delta}(n)$ to be $\epsilon^2$ plus an upper bound on estimation error which goes to zero with $n$ (note that one can still get sharp $\widetilde{O}(n^{-p})$-type estimation error bounds via offline regression in the misspecified setting; see \citealt{rakhlin2017empirical}). Plugging the above choice of $\cE_{\cF,\delta}(n)$ into Algorithm \ref{alg:falcon+} and the general regret bound (\ref{eq:bound}), one can easily obtain the regret bound in the misspecified setting, which typically equals to the regret bound in the well-specified setting plus an additive term of $O(\epsilon\sqrt{K}T)$. %
While this additive term is linear in $T$, it is not surprising and is consistent with existing results in such a setting (e.g., Theorem 5 of \citealt{foster2020beyond}), as the model is misspecified while the regret is still evaluated against the globally optimal policy $\pi_{f^*}$. %

It is worth noting that the misspecification error $\epsilon$ may be \emph{unknown} in practice. %
The challenge of adapting to an unknown $\epsilon$  is addressed by follow-up work of our paper; see \S\ref{sec:conclusion} for a discussion of follow-up work.

\textbf{Rewards can be unbounded/heavy-tailed.} We note that the  assumption $r_t\in[0,1]^{\cA}$ is not essential to our reduction, if we only want to bound the regret \emph{in expectation} rather than with high probability. Specifically, to obtain the same results on the expected regret, we only need the following condition on the reward distribution:
\begin{equation}\label{eq:condition}\E_{x\sim\cD}[\sup_{a,a'\in\cA}(f^*(x,a)-f^*(x,a'))]\le\sqrt{K},\end{equation}
which is very weak --- in the special case of multi-armed bandits, it means ``the gap between the \emph{mean rewards} of two actions is no greater than $\sqrt{K}$.'' Note that (\ref{eq:condition}) only concerns the conditional mean $f^*(x,a)$ rather than the reward distribution, thus allows the ``random noise'' in the reward to have an \emph{arbitrary} distribution. Moreover, (\ref{eq:condition}) allows the scale of $f^*(x,a)$ to be arbitrarily large and unknown (as it only concerns the gap), thus enables $\cF$ to contain unbounded functions. As a consequence, recent advances on ``fast rates'' for offline unbounded/heavy-tailed regression (see \citealt{mendelson2014learning} and Section 8 of \citealt{xu2020towards}) can be  translated into contextual bandits. Here, the merit of our reduction is that different assumptions on the reward distribution only affect our results through the offline learning guarantee  $\cE_{\cF,\delta}(n)$ in Assumption \ref{assump:oracle}, thus the associated offline regression challenges are ``separated'' from  contextual bandits. Note that while heavy-tailed noise is very well-studied in offline regression, it is rarely studied in contextual bandits, especially with general function classes. Our reduction provides a simple way to obtain such results. %

\textbf{Robustness to delayed and batched feedback.}  In practical applications of contextual bandits (e.g., clinical trials, recommendation systems), the feedback to the learner is typically not immediate and may arrive in batches (\citealt{chapelle2011empirical}). In \cite{perchet2016batched} and \cite{gao2019batched}, a ``batched bandit'' model is developed, where the the learner must split her learning process into a small number of batches due to several practical constraints.  
Since \texttt{FALCON} / \texttt{FALCON+} only requires processing the feedback associated with each epoch after this epoch ends, our algorithm naturally handles delayed and batched feedback. In particular, Theorem \ref{thm:inf} is directly applicable to the batched version of stochastic contextual bandits with general function classes, and implies that $O(\log\log T)$  batches are sufficient for one to achieve the optimal regret rate of $T$, which significantly generalizes existing results on batched bandits. We note that the ability to handle delayed and batched feedback is an important advantage of adopting an epoch schedule and using an offline regression oracle rather than an online regression oracle, as online regression is known to require immediate feedback, and the increase of regret due to  delayed rewards is generally much larger in adversarial models than in stochastic models (see \citealt{lattimore2018bandit}).

\section{Regret Analysis}\label{sec:analysis}
In this section, we elaborate on how our simple algorithm achieves the optimal regret. While we present our analysis based on Algorithm \ref{alg:cbor} and Theorem \ref{thm:finite}, everything is essentially the same for Algorithm \ref{alg:falcon+} and Theorem \ref{thm:inf}. 
We first analyze our algorithm (through an interesting dual interpretation) and provide  in \S\ref{sec:tale} to \S\ref{sec:ocbl} a proof sketch of Theorem \ref{thm:finite}. Then, in \S\ref{sec:bypass}, we explain the key idea behind our algorithm, and in \S\ref{sec:born}, we show how this idea leads to the algorithm.

For ease of presentation, in this section we assume that $|\cX|<\infty$ but allows $|\cX|$ to be arbitrarily large. Focusing on such a setting enables us to highlight important ideas and key insights without the need to invoke measure theoretic arguments (which are necessary for infinite/uncountable $\cX$). We remark that all our results hold for general uncountable $\cX$; see Appendix \ref{app:measure} for more details.

Since some notations appearing in Algorithm \ref{alg:cbor} are shorthand and do not explicitly reveal the dependence between different quantities (e.g.,  $\widehat{a}_t$ and $p_t(\cdot)$ should be written as a function and a conditional distribution explicitly depending on the random context $x_t$), we introduce some new notations which can describe the decision generating process of Algorithm \ref{alg:cbor} in a more systematic way. For each epoch $m\in\N$, given the learning rate $\gamma_m\in\mathbb{R}_+$ and the greedy predictor $\widehat{f}_m:\cX\times\cA\rightarrow[0,1]$ (which are uniquely determined by the data from the first $m-1$ epochs), we can explicitly represent the algorithm's decision rule using $\gamma_m$ and $\widehat{f}_m$. Specifically, for any $x\in\cX$, define 
$\widehat{a}_m(x):=\max_{a\in\cA}\widehat{f}_m(x,a)$ and
\[
p_m(a\mid x):=\begin{cases}\frac{1}{K+\gamma_m\left(\widehat{f}_m(x,\widehat{a}_m(x))-\widehat{f}_m(x,a)\right)}, &\text{for all } a\ne \widehat{a}_m(x),\\
1-\sum_{a\ne \widehat{a}_m(x)}p_m(a\mid x), &\text{for }a=\widehat{a}_m(x).
\end{cases}
\]
Then  $p_m(\cdot\mid\cdot)$ is a well-defined \emph{action selection kernel} (see \S\ref{sec:notation})
that completely characterizes the algorithm's decision rule in epoch $m$. Specifically, at each round $t$ in epoch $m$, the algorithm first observes a random context $x_t$, then samples its action $a_t$ according to the conditional distribution $p_m(\cdot\mid x_t)$. %
Note that $p_m(\cdot\mid\cdot)$ depends on all the randomness up to round $\tau_{m-1}$ (including round $\tau_{m-1}$), which means that $p_m(\cdot\mid\cdot)$ depends on $p_1(\cdot\mid\cdot),p_2(\cdot\mid\cdot),\dots,p_{m-1}(\cdot\mid\cdot)$, and will affect $p_{m+1}(\cdot\mid\cdot),p_{m+2}(\cdot\mid\cdot),\dots$ in later epochs.

\subsection{A Tale of Two Processes}\label{sec:tale}

The conventional way of analyzing our algorithm's behavior at round $t$ in epoch $m$ is to study the following \textit{original} process:
\begin{enumerate}
    \item Nature generates $x_t\sim\cD$.
    \item Algorithm samples $a_t\sim p_t(\cdot)$.
\end{enumerate}
The above process is however tricky to analyze, because the algorithm's sampling strategy over actions, $p_t(\cdot)=p_m(\cdot\mid x_t)$, \text{depends on} the new random context $x_t$, and cannot be evaluated in advance before observing $x_t$. %

A core idea of our analysis is to find a way to examine the algorithm's behavior at round $t$ before observing $x_t$. To this end, we look at the following \textit{virtual} process at round $t$ in epoch $m$:
\begin{enumerate}
    \item Algorithm samples $\pi_t\sim Q_m(\cdot)$, where $\pi_t:\cX\rightarrow\cA$ is a \textit{policy}, and $Q_m(\cdot): \cA^{\cX}\rightarrow[0,1]$ is a probability distribution over all policies in $\cA^{\cX}$.
    \item Nature generates $x_t\sim\cD$.
    \item Algorithm selects $a_t=\pi_t(x_t)$ deterministically.  
\end{enumerate}
The merit of the above process is that the algorithm's sampling procedure over policies, $Q_m(\cdot)$, is \text{independent of} the new context $x_t$. While the algorithm still has to select an action based on $x_t$ in step 3, this is completely deterministic and easier to analyze. Note that at round $t$,  $Q_m(\cdot)$ is a stationary distribution which has already been determined at the beginning of epoch $m$.

The second process is however a \textit{virtual} process because it is not how our algorithm directly proceeds. An immediate question is whether we can always find a distribution over policies $Q_m(\cdot)$, such that our algorithm behaves exactly the same as the \text{virtual} process in epoch $m$?
Recall that the algorithm's decision rule in epoch $m$ is completely characterized by the action selection kernel $p_m(\cdot\mid\cdot)$. In fact, any action selection kernel $p_m(\cdot\mid\cdot)$ can be translated into an ``equivalent'' distribution over policies $Q_m(\cdot)$, enabling us to  study our algorithm's behavior through the virtual process. We complete this translation in \S\ref{sec:translation}.

\subsection{Action Selection Kernel as a Randomized Policy}\label{sec:translation}

We define the \textit{universal policy space} as $\Psi:={\cA}^{\cX}$, 
which contains all possible policies. For any $p_m(\cdot\mid\cdot)$, 
we can construct a (unique)  product probability measure $Q_m(\cdot)$ on $\Psi$ such that
$
Q_m(\pi)=\prod_{x\in\cX}p_m(\pi(x)\mid x)
$ for all $\pi\in\Psi$ (see Lemma \ref{lm:map} in the appendix). 
This $Q_m(\cdot)$ 
ensures that for every $x\in\cX,a\in\cA$, %
\begin{equation}\label{eq:translation}
p_m(a\mid x)=\sum_{\pi\in\Psi}\1\{\pi(x)=a\}Q_m(\pi).
\end{equation}
That is, for any arbitrary context $x$, the algorithm's action generated by $p_m(\cdot\mid x)$ is probabilistically equivalent to the action generated by $Q_m(\cdot)$  through the virtual process in \S\ref{sec:tale}. Since $Q_m(\cdot)$ is a dense distribution over all \textit{deterministic} policies in the universal policy space, we refer to $Q_m(\cdot)$ as the ``equivalent \textit{randomized} policy'' induced by $p_m(\cdot\mid\cdot)$. %
Since $p_m(\cdot\mid\cdot)$ is completely determined by $\gamma_m$ and $\hat{f}_m$, we know that $Q_m(\cdot)$ is also completely determined by $\gamma_m$ and $\hat{f}_m$.

We emphasize that our algorithm does not compute $Q_m(\cdot)$, but implicitly maintains $Q_m(\cdot)$ through $\gamma_m$ and $\widehat{f}_m$. This is important, as even when $\cX$ is known to the learner, %
computing the product measure $Q_m(\cdot)$ requires $\Omega(|\cX|)$ computational cost which is intractable for large $|\cX|$. Remember that all of our arguments based on $Q_m(\cdot)$ are only applied for the purpose of statistical analysis and have nothing to do with the algorithm's original implementation.

\subsection{Dual Interpretation in the Universal Policy Space}\label{sec:dualnew}
Through the lens of the virtual process, we find a {dual interpretation} of our algorithm: \textit{it sequentially maintains a dense distribution $Q_m(\cdot)$ over all the policies in the universal policy space $\Psi$, for epoch $m=1,2,3\dots$}. The analysis of the behavior of our algorithm thus could  hopefully reduce to the analysis of an evolving sequence $\{Q_m\}_{m\in\N}$ (which is still non-trivial because it still depends on all the interactive data). All our analysis from now on will be based on the above dual interpretation.

As we start to explore how $\{Q_m\}_{m\in\N}$ evolves in the  universal policy space, let us first define some \textit{implicit} quantities in this world which are useful for our statistical analysis --- they are called ``{implicit}'' because our algorithm does not really compute or estimate them at all, yet they are all well-defined and implicitly exist as long as our algorithm proceeds.

Define the ``implicit reward'' of a policy $\pi\in\Psi$ as
$$
\cR(\pi):=\E_{x\sim\cD}\left[f^*(x,\pi(x))\right]
$$
and define the ``implicit regret''\footnote{Note that this is an ``instantaneous'' quantity in $[0,1]$, not a sum over multiple rounds.} of a policy $\pi\in\Psi$ as
$$
\Reg(\pi):=\cR(\pi_{f^*})-\cR(\pi).
$$
At round $t$ in epoch $m$, given a predictor $\widehat{f}_m$, define the ``predicted implicit reward'' of a policy $\pi\in\Psi$ as
$$
\wcR_t(\pi):=\E_{x\sim\cD}\left[\widehat{f}_{m}(x,\pi(x))\right]
$$
and define the ``predicted implicit regret'' of a policy $\pi\in\Psi$ as\footnote{Note that in \S\ref{sec:setting} we have defined $\pi_f$ as the reward-maximizing policy induced by a reward function $f$, i.e., $\pi_f(x)=\arg\max_{a\in\cA}f(x,a)$ for all $x\in\cX$. Also note that  \textit{not all} policies in $\Psi$ can be written as $\pi_f$ for some $f\in\cF$.}
$$
\wReg_t(\pi):=\wcR_t(\pi_{\widehat{f}_{m}})-\wcR_t(\pi).
$$
The idea of defining the above quantities is motivated by the celebrated work of \cite{agarwal2014taming}, which studies policy-based optimal contextual bandit learning in the agnostic setting (in which setting the above quantities are not implicit but play obvious roles and are directly estimated by their algorithm). There are some differences in the definitions though. For example,  \cite{agarwal2014taming} define the above quantities for all policies $\pi$ in a given finite policy class $\Pi$, while we define the above quantities for all policies in the universal policy space $\Psi$ (which is strictly larger than $\Pi$). Also,  \cite{agarwal2014taming} define $\wcR_t(\pi)$ and $\wReg_t(\pi)$ based on the inverse propensity scoring estimates, while we define them based on a single predictor.
We will revisit these differences later.

After defining the above quantities, we make a simple yet powerful observation, which is an immediate consequence of (\ref{eq:translation}): for any epoch $m\in\N$ and any round $t$ in epoch $m$, we have
$$
\E_{(x_t,r_t)\sim\mcD,a_t\sim p_m(\cdot\mid x_t)}\left[r_t(\pi_{f^*})-r_t(a_t)\mid \gamma_m, \widehat{f}_m\right]=\sum_{\pi\in\Psi}Q_m(\pi)\Reg(\pi),
$$
see Lemma \ref{lm:equi} in the appendix. This means that (under any possible realization of $\gamma_m,\widehat{f}_m$) the expected instantaneous regret incurred by our algorithm  equals to the ``implicit regret'' of the randomized policy $Q_m$ (as a weighted sum over the implicit regret of every deterministic policy $\pi\in\Psi$). Since $\Reg(\pi)$ is a fixed deterministic quantity for each $\pi\in\Psi$, the above equation indicates that to analyze our algorithm's expected regret in epoch $m$, we only need to analyze the distribution $Q_m(\cdot)$. This property shows the advantage of our dual interpretation: compared with the original process in \S\ref{sec:tale} where it is hard to evaluate our algorithm without $x_t$, now we can evaluate our algorithm's behavior regardless of $x_t$.

\subsection{Optimal Contextual Bandit Learning in the Universal Policy Space}\label{sec:ocbl}
We proceed to understand how $Q_m(\cdot)$ evolves in the universal policy space. We first state an immediate observation based on the equivalence of $p_m(\cdot\mid\cdot)$ and $Q_m(\cdot)$ given by equation (\ref{eq:translation}).

\begin{observation}
For any deterministic policy $\pi\in\Psi$, the quantity $\E_{x\sim\cD}\left[{\frac{1}{p_m(\pi(x)\mid x)}}\right]$ is the expected inverse probability that  ``the decision generated by the randomized policy $Q_m$ is the same as the decision generated by the deterministic policy $\pi$,'' over the randomization of context $x$. This quantity can be intuitively understood as a  measure of the ``decisional divergence'' between the randomized policy $Q_m$ and the deterministic policy $\pi$.
\end{observation}

Now let us utilize the closed-form structure of $p_m(\cdot\mid x)$ in our algorithm and point out a most important property of $Q_m(\cdot)$ stated below (see Lemma \ref{lm:op1} and Lemma \ref{lm:op2} in the appendix for details).

\begin{observation}\label{obs:2}
For any epoch $m\in\N$ and any round $t$ in epoch $m$, for any possible realization of $\gamma_m$ and $\widehat{f}_m$, $Q_m(\cdot)$ is a feasible solution to the following ``{Implicit Optimization Problem}'' (\texttt{IOP}):
\begin{align}
    &\sum_{\pi\in\Psi}Q_m(\pi)\wReg_{t}(\pi)\le {K}/{\gamma_m},\label{eq:op11}\\
    \forall \pi\in\Psi,~~~&\E_{x\sim\cD}\left[\frac{1}{p_m(\pi(x)\mid x)}\right]\le K+\gamma_m\wReg_{t}(\pi).\label{eq:op22}
\end{align}
\end{observation}

We give some interpretations for the ``{Implicit Optimization Problem}'' (\texttt{IOP}) defined above. (\ref{eq:op11}) says that  $Q_m$ controls its  predicted implicit regret (as a weighted sum over the predicted implicit regret of every policy $\pi\in\Psi$, based on the predictor $\widehat{f}_m$) within $K/\gamma_m$. This can be understood as an ``exploitation constraint'' because it require $Q_m$ to put more mass on “good policies” with low predicted implicit regret (as judged by the current predictor $\widehat{f}_m$). (\ref{eq:op22}) says that the decisional divergence between $Q_m(\cdot)$ and any policy $\pi\in\Psi$ is controlled by the predicted implicit regret of policy $\pi$ (times a learning rate $\gamma_m$ and plus a constant $K$). This can be understood as an ``adaptive exploration constraint,'' as it requires that $Q_m$  behaves similarly to \textit{every} policy $\pi\in\Psi$ at some level (which means that there should be sufficient exploration), while allowing $Q_m$ to be more similar to ``good policies'' with low predicted implicit regret and less similar to ``bad policies'' with high predicted implicit regret (which means that the exploration can be conducted adaptively based on the judgement of the predictor $\widehat{f}_m$). Combining (\ref{eq:op11}) and (\ref{eq:op22}), we conclude that $Q_m$ elegantly strikes a balance between exploration and exploitation --- it is surprising that this is done completely implicitly, as the original algorithm does not explicitly consider these constraints at all.

There are still a few important tasks to complete. The first task is to figure out what exactly the decisional divergence $\E_{x\sim\cD}\left[{\frac{1}{p_m(\pi(x)\mid x)}}\right]$ means. We give an answer in Lemma \ref{lm:reward-estimates}, which shows that with high probability, for any epoch $m\in\N$ and any round $t$ in epoch $m$, for all $\pi\in\Psi$,
$$
|\wcR_t(\pi)-\cR(\pi)|\le\frac{\sqrt{K}}{2\gamma_{m}}\sqrt{\max_{1\le n \le m-1}\E_{x\sim\cD}\left[{\frac{1}{p_n(\pi(x)\mid x)}}\right]}.
$$
That is, the prediction error of the implicit reward of every policy $\pi\in\Psi$ can be bounded by the (maximum) decisional divergence between $\pi$ and all previously used randomized policies $Q_1,\dots,Q_{m-1}$. This is consistent with our intuition, as the more similar a policy is to the previously used randomized policies, the more likely that this policy is implicitly explored in the past, and thus the more accurate our prediction on this policy should be. We emphasize that the above inequality  relies on our specification of the learning rate $\gamma_m$: we can bound the prediction error using $1/\gamma_m$ because $1/\gamma_m$ is proportional to $1/\sqrt{\tau_{m-1}}$ and proportional to $\sqrt{\log|\cF|}$ --- the first quantity $1/\sqrt{\tau_{m-1}}$ is related to the length of the history, and the second quantity $\sqrt{\log|\cF|}$ is related to the generalization ability of function class $\cF$. This is the first place that our proof requires an epoch-varying learning rate.

The second task is to further bound (the order of) the prediction error of the implicit regret of every policy $\pi$, as the implicit regret is an important quantity that can be directly used to bound our algorithm's expected regret (see \S\ref{sec:dualnew}). We do this in Lemma \ref{lm:regret-estimates}, where we show that with high probability, for any epoch $m\in\N$ and any round $t$ in epoch $m$, for all $\pi\in\Psi$,
\[
\Reg(\pi)\le 2\wReg_t(\pi)+5.15K/\gamma_{m},
\]
\[
\wReg_t(\pi)\le 2\Reg(\pi)+5.15K/\gamma_{m}
\]
through an inductive argument. While this is a uniform-convergence-type result, we would like to clarify that this does not mean that there is a uniform convergence of $|\Reg(\pi)-\wReg_t(\pi)|$ for all $\pi\in\Psi$, which is too strong and unlikely to be true. Instead, we use a smart design of $\Reg(\pi)-2\wReg_t(\pi)$ and $\wReg_t(\pi)-2\Reg(\pi)$ (the design is motivated by Lemma 13 in \citealt{agarwal2014taming}), which enables us to characterize the fact that the predicted implicit regret of ``good policies'' are becoming more and more accurate, while the predicted implicit regret of ``bad policies'' do not need to be accurate (as their orders directly dominate $K/\gamma_m$). We emphasize that the above result relies on the fact that our learning rate $\gamma_m$ is gradually increasing from $O(1)$ to $O(\sqrt{T})$, as we use an inductive argument and in order to let the hypothesis hold for initial cases we have to let $\gamma_m$ be very small for small $m$. This is the second place that our proof requires a epoch-varying learning rate.

We have elaborated on how our algorithm implicitly strikes a balance between exploration and exploitation, and how our algorithm implicitly enables some nice uniform-convergence-type results to happen in the universal policy space. This is already enough to guarantee that the dual interpretation of our algorithm achieves optimal contextual bandit learning in the universal policy space. The rest of the proof is standard and can be found in the appendix.

\subsection{Key Idea: Bypassing the Monster}\label{sec:bypass}
For readers who are familiar with the research line of optimal contextual bandits learning in the agnostic setting using an offline cost-sensitive classification oracle (represented by \citealt{dudik2011efficient,agarwal2014taming}), they may find a surprising connection between the \texttt{IOP} (\ref{eq:op11}) (\ref{eq:op22}) that we introduce in Observation \ref{obs:2} and the so-called ``Optimization Problem'' (\texttt{OP}) in \cite{dudik2011efficient} and \cite{agarwal2014taming} --- in particular, if one takes a look at the \texttt{OP} defined in page 5 of \cite{agarwal2014taming}, one will find that it is almost the same as our \texttt{IOP} (\ref{eq:op11}) (\ref{eq:op22}), except for two fundamental differences:
\begin{enumerate}
    \item The \texttt{OP} of \cite{dudik2011efficient} and \cite{agarwal2014taming} is defined on a given finite policy class $\Pi$, which may have an arbitrary shape. As a result, to get a solution to \texttt{OP}, the algorithm must explicitly solve a complicated (non-convex) optimization problem over a possibly complicated policy class --- this requires a considerable number of calls to a cost-sensitive classification oracle, and is the major computational burden of \cite{dudik2011efficient} and \cite{agarwal2014taming}. Although \cite{agarwal2014taming} ``tame the monster'' and reduce the computational cost by only strategically maintaining a \textit{sparse} distribution over policies in $\Pi$, solving \texttt{OP} still requires $\widetilde{O}(\sqrt{KT/\log|\Pi|})$ calls to the classification oracle and is computationally expensive --- the monster is still there.
    
    By contrast, our \texttt{IOP} is defined on the universal policy space $\Psi$, which is a nice product space. The \texttt{IOP} can thus be viewed as a very ``slack'' relaxation of \texttt{OP} which is extremely easy to solve. In particular, as \S\ref{sec:analysis} suggests, the solution to \texttt{IOP} can have a completely decomposed form which enables our algorithm to solve it in a complete \textit{implicit} way. This means that our algorithm can  implicitly and  confidently maintain a \textit{dense} distribution over all policies in $\Psi$, while solving \texttt{IOP} in closed forms at no computational cost --- there is no monster any more as we simply bypass it.

    \item In \cite{dudik2011efficient} and \cite{agarwal2014taming}, the quantities $\wcR_t(\pi)$ and $\wReg_t(\pi)$ are explicitly  calculated based on the model-free inverse propensity scoring estimates. As a result, their regret guarantees do not require the realizability assumption.
    
    By contrast, in our paper, the quantities $\wcR_t(\pi)$ and $\wReg_t(\pi)$ are implicitly calculated based on a single greedy predictor $\widehat{f}$ ---  we can do this because we have the realizability assumption (or relaxed notions of realizability) which enables us to \emph{learn the reward model} and obtain an  $\widehat{f}$ that is close to $f^*$ (see also Assumption \ref{assump:oracle}). As a result, we  make a single call to the offline regression oracle here, and this is the main computational cost of our algorithm.
\end{enumerate}

A possible question could then be that, given the fact that the main computational burden of \cite{dudik2011efficient} and \cite{agarwal2014taming} is solving \texttt{OP}, why can't they simply relax \texttt{OP} as we do in our \texttt{IOP}? The answer is that without the realizability assumption, they have to rely on the capacity control of their policy space, i.e., the boundedness of  $|\Pi|$, to obtain their statistical guarantees. Indeed, as their $\widetilde{O}(\sqrt{KT\log|\Pi|})$ regret bound suggests, if one let $\Pi=\cA^\cX$, then the regret could be as large as $\Omega(|\cX|)$. Specifically, their analysis requires the limited capacity (or complexity) of $\Pi$ in two places: first, a generalization guarantee of the inverse propensity scoring requires limited $|\Pi|$; second, since they have to explicitly compute $\wcR_t(\pi)$ and $\wReg_t(\pi)$ without knowing the true context distribution $\cD$, they try to  approximate it based on the historical data, which also requires limited $|\Pi|$ to enable statistical guarantees.

Our algorithm bypasses the above two requirements simultaneously: first, since we use model-based regression rather than model-free inverse propensity scoring to make predictions, we do not care about the complexity of our policy space in terms of prediction (i.e., the generalization guarantee of our algorithm is governed by the capacity of $\cF$ rather than $\Psi$); second, since our algorithm does not require explicit computation of $\wcR_t(\pi)$ and $\wReg_t(\pi)$, we do not care about what $\cD$ looks like. Essentially, all of these nice properties originate from the realizability assumption. This is how we understand the value of realizability: it does not only (statistically) give us better predictions, but also (computationally) enables us to remove the restrictions in the policy space , which helps us to bypass the monster. 

\subsection{The Birth of \texttt{FALCON}}\label{sec:born}
The idea behind ``bypassing the monster,'' as explained in \S\ref{sec:bypass}, is exactly what leads to the derivation of the \texttt{FALCON} algorithm. The  derivation is interesting because it reveals deep connections between the celebrated \texttt{OP} studied by \cite{dudik2011efficient,agarwal2014taming} and the intriguing probabilistic selection strategy studied by \cite{abe1999associative} and \cite{foster2020beyond}. Before we close this section, we describe how \texttt{FALCON} was derived. We hope that this derivation process can provide new perspectives on previous work, and motivate further discovery of new algorithms for other bandit and reinforcement learning problems.

\begin{enumerate}
    \item We conduct a thought experiment, considering how \texttt{ILOVETOCONBANDITS} (\citealt{agarwal2014taming}) can solve our problem without the realizability assumption, given an induced  policy class $\Pi=\{\pi_f\mid f\in\cF\}$.
    \item \texttt{ILOVETOCONBANDITS} uses an inverse propensity scoring approach to calculate the predicted reward and predicted regret of policies. This can be thought as using a model-free approach (different from our \S\ref{sec:dualnew}) to calculate $\wcR_t(\pi)$ and $\wReg_t(\pi)$ for $\pi\in\Pi$.
    \item The computational burden in the above thought experiment is to solve \texttt{OP} over $\Pi$, which requires repeated calls to a cost-sensitive classification oracle.
    \item When we have realizability, we can use a regression oracle to obtain a predictor $\widehat{f}_m$ and use it to calculate $\wcR_t(\pi)$ and $\wReg_t(\pi)$ for $\pi\in\Psi$ (if $\cD$ is known).  Here, we can operate on the set of all policies rather than only on $\Pi$, as generalization is governed by the capacity of $\cF$.
    \item An early technical result of Lemma 4.3 in \cite{agarwal2012contextual} is very interesting. It shows that when one tries to solve contextual bandits using regression approaches, one should try to bound a quantity like ``the expected inverse probability of choosing the same action'' --- note that a very similar quantity also appears in \texttt{OP} in \cite{agarwal2014taming}. This suggests that an offline-regression-oracle-based algorithm should try to satisfy some requirements similar to \texttt{OP}. (Lemma 4.3 in \cite{agarwal2012contextual} also motivates our Lemma \ref{lm:reward-estimates}. But our Lemma \ref{lm:reward-estimates} goes a significant step beyond Lemma 4.3 in \cite{agarwal2012contextual} 
    by unbinding the relationship between a predictor and a policy and moving forward to the universal policy space.)
    \item Motivated by 3, 4, and 5, we relax the domain of \texttt{OP} from $\Pi$ to $\Psi$, and obtain the relaxed problem \texttt{IOP}. Since the new domain $\Psi=\cA^\cX$ is a product space, we consider the per-context decomposed version of \texttt{IOP}, i.e., a problem ``conditional on a single $x$'':
    \begin{align*}
    \sum_{\pi(x)\in\cA}p_m(\pi(x)\mid x){\gamma_m}\left(\widehat{f}_m(\pi_{\widehat{f}_m}(x))-\widehat{f}_m(\pi(x))\right)\le {K},\\
    \forall \pi(x)\in\cA,~~~\frac{1}{p_m(\pi(x)\mid x)}\le K+\gamma_m\left(\widehat{f}_m(\pi_{\widehat{f}_m}(x))-\widehat{f}_m(\pi(x))\right).
\end{align*}
    Clearly, there is a closed-form solution to the above problem: the conditional probability of selecting an action $\pi(x)$ should be inversely proportional to the predicted reward gap of $\pi(x)$ times $\gamma_m$. This leads to \texttt{FALCON}'s decision generating process in epoch $m$.
\end{enumerate}

\section{Concluding Remarks}\label{sec:conclusion}
In this paper, we propose the first provably optimal offline-regression-oracle-based algorithm for general contextual bandits, solving an important open problem in the contextual bandit literature. Our algorithm is surprisingly fast and simple, and our analysis is quite general. We hope that our findings can motivate future research on contextual bandits and reinforcement learning. We discuss some follow-up work and future directions below.

\textbf{Follow-up work.} Since the first version of our paper appeared on arXiv (\citealt{simchi2020bypassing}), there have
been several developments directly inspired by our work. Here we mention several extensions of our results. 
\cite{xu2020upper} extend our results to the practical setting of infinite actions.  \cite{foster2020instance} build on our results to achieve instance-dependent guarantees of contextual bandits, and further extend the results to reinforcement learning. \cite{wei2021non} extend our results to non-stationary contextual bandits; their approach to deal with non-stationarity is quite general and finds broader applications. 
\cite{sen2021top} extend our results to a combinatorial action model where one need to select more than one action per round. \cite{krishnamurthy2021adapting} extend our results to the setting where the model is misspecified and the misspecification error is unknown.

\textbf{Future directions.} Going forward, our work motivates many interesting research questions. First, in Example \ref{ex:linear} (linear contextual bandits), our regret bound has worse dependence on $K$ compared with \texttt{LinUCB} (\citealt{chu2011contextual}). This seems like a limitation of the employed probabilistic selection strategy, i.e., it does not fully utilize the special properties of some function classes to obtain improved dependence on $K$. Understanding this issue better, and more broadly, understanding how to characterize and achieve the regret's optimal dependence on $K$ for general function classes, is important from both theoretical and practical points of view.  Second, our work establishes new connections between policy-based (agnostic) and value-function-based (realizable) contextual bandits. %
We hope that the  techniques and perspectives developed in this paper can find broader applications in reinforcement learning with function approximation. Finally, our work successfully reduces a prominent online decision making problem to a well-studied offline supervised learning problem. Can similar online-to-offline reductions be achieved in other practical learning settings? %

\section*{Acknowledgments}
The authors would like to thank the review team for their helpful comments; in particular, for pointing out  some observations in \S\ref{sec:advantages}. 
 	The authors would like to express sincere gratitude to Alekh Agarwal, Dylan Foster, Akshay Krishnamurthy, John Langford, Menglong Li, Alexander Rakhlin, Yining Wang and Yunbei Xu for helpful comments and discussions. The authors would like to thank the support from the MIT-IBM partnership in AI.

\bibliographystyle{informs2014} %
\bibliography{bibliography} %

\begin{appendices}

\section{Proof of Theorem \ref{thm:finite} and Theorem \ref{thm:inf}}\label{app:thm1}\label{app:thm2}
Since Theorem \ref{thm:inf} is almost a strict generalization of Theorem \ref{thm:finite} (and Corollary \ref{cor:finite}), we prove them together by conducting a unified analysis. To ensure that the notations are compatible, in some lemmas we will state our results under two different setups, each of which provides consistent definitions of $(\widehat{f}_m)_{m\in\N}$ and $(\gamma_m)_{m\in\N}$ and makes consistent assumptions.
\begin{setup}\label{setup:1}
We consider the learning model in \S\ref{sec:alg}, where $|\cF|<\infty$, $f^*\in\cF$,
    and all rewards are $[0,1]$-bounded. %
    In this setup, $(\widehat{f}_m)_{m\in\N}$ and $(\gamma_m)_{m\in\N}$ are given by Algorithm \ref{alg:cbor} with $c=1/30$.
\end{setup}
\begin{setup}\label{setup:2}
We consider the learning model in \S\ref{sec:extension} and do not make any assumption on $\cF$, $f^*$, or the reward distribution. We only assume Assumption \ref{assump:oracle} and condition (\ref{eq:condition}) for now (as a result, we allow $f^*\notin\cF$, and allow rewards to be unbounded/heavy-tailed). In this setup, $(\widehat{f}_m)_{m\in\N}$ and $(\gamma_m)_{m\in\N}$ are given by Algorithm \ref{alg:falcon+} with $c=1/2$. Moreover, we assume that $\tau_m\ge2^m$ for $m\in\N$; for such epoch schedules, it is essentially without loss of generality to assume that $(\gamma_m)_{m\in\N}$ are non-decreasing.
\end{setup}

As we can see, Setup \ref{setup:2} involves a much more general learning model. However, Setup \ref{setup:1} allows $(\widehat{f}_m)_{m\in\N}$ to be generated with data reuse (i.e., one can feed the data collected in all previous epochs into the offline regression oracle) and 
does not require the epoch length to grow geometrically. One can understand Setup \ref{setup:1} as an example of utilizing martingale concentration results (Lemma \ref{lm:ag12-2}) obtained  under additional model assumptions to show some additional properties.

Before we start our proof of Theorem \ref{thm:finite} (under Setup \ref{setup:1}) and Theorem \ref{thm:inf} (under Setup \ref{setup:2}), we make an important remark regarding our presentation. In the main part of this appendix (Appendix \ref{app:def} to Appendix \ref{app:final}), we give a detailed proof (of both theorems) which closely follows the proof sketch in Section \ref{sec:analysis}, \emph{under the condition that  $|\cX|<\infty$}. Such a condition enables us to give an insightful proof without the need to worry about measurability issues. However, when $\cX$ is infinite or (more generally) uncountable, one has to deal with 
the measurability issues arising from the presence of uncountable probability spaces. While rigorous discussions of measurability issues are usually omitted in the contextual bandit literature (for brevity or for simplicity), we feel that a discussion of such issues is important here due to our extensive use of the \emph{universal policy space} $\cA^{\cX}$, which would easily contain non-measurable policies when $\cX$ is uncountable. Therefore, in the last part of this appendix (Appendix \ref{app:measure}), we consider general uncountable $\cX$ and present a simple fix to the associated measurability issues, showing that our results are indeed general.

\subsection{Definitions}\label{app:def}
For notational convenience, we make some definitions. Some of the definitions have appeared in the main article. For all $t=1,\dots,T$, we let $\Upsilon_t:=\sigma((x_1,r_1,a_1),\cdots,(x_t,r_t,a_t))$ denote the sigma-algebra generated by the history up to round $t$ (inclusive), and let $m(t):=\min\{m\in\N: t \le \tau_m\}$ denote the epoch that round $t$ belongs to. Let $\Psi:=\cA^\cX$ be the \emph{universal policy space} (which is defined via taking the \emph{Cartesian product}). For any action selection kernel $p$ and any policy $\pi\in\Psi$, define
\[
V(p,\pi):=\E_{x\sim\cD}\left[\frac{1}{p(\pi(x)\mid x)}\right],
\]
\[
\cV_t(\pi):=\max_{1\le m \le m(t)-1}\{V(p_m,\pi)\}.
\]
For any $\pi\in\Psi$, define
\[
\cR(\pi):=\E_{x\sim\cD}\left[f^*(x,\pi(x))\right],
\]
\[
\wcR_t(\pi):=\E_{x\sim\cD}\left[\widehat{f}_{m(t)}(x,\pi(x))\right],
\]
\[
\Reg(\pi):=\cR(\pi_{f^*})-\cR(\pi),
\]
\[
\wReg_t(\pi):=\wcR_t(\pi_{\widehat{f}_{m(t)}})-\wcR_t(\pi).
\]

\subsection{High-probability Events}\label{app:hpe}

Lemma \ref{lm:ag12-2} and Lemma \ref{lm:oracle} present some basic concentration results. %

\begin{lemma}\label{lm:ag12-2} Consider Setup \ref{setup:1}. For all $m\ge2$, with probability at least $1-{\delta}/{(2m^2)}$,
we have:
\begin{align*}
&~~~\sum_{t=1}^{\tau_{m-1}}\E_{x_t,a_t}\left[(\widehat{f}_{m}(x_t,a_t)-f^*(x_t,a_t))^2\mid\Upsilon_{t-1}\right]\\
&=\sum_{t=1}^{\tau_{m-1}}\E_{x_t,r_t,a_t}\left[(\widehat{f}_{m}(x_t,a_t)-r_t(a_t))^2-(f^*(x_t,a_t)-r_t(a_t))^2 \mid \Upsilon_{t-1}\right]\\
&\le 100\log\left(\frac{2|\cF|m^2\log_2(\tau_{m-1})}{\delta}\right)\le225 \log\left(\frac{|\cF|m\log(\tau_{m-1})}{\delta}\right)=\frac{K}{4\gamma_m^2}.
\end{align*}
Therefore (by a union bound), the following event $\Gamma_1$ holds with probability at least $1-\delta/2$:
\begin{equation*}\label{eq:clean-event}
\Gamma_1:=\left\{\forall m\ge2,~~\frac{1}{\tau_{m-1}}\sum_{t=1}^{\tau_{m-1}}\E_{x_t,a_t}\left[(\widehat{f}_{m}(x_t,a_t)-f^*(x_t,a_t))^2\mid\Upsilon_{t-1}\right]\le \frac{K}{4\gamma_m^2}\right\}.
\end{equation*}
\end{lemma}

Lemma \ref{lm:ag12-2} follows from Lemma 4.1 (see equation (4.1) therein) and Lemma 4.2 in \cite{agarwal2012contextual}, and we omit the proof here (compared to their proof, we just slightly change the way of taking union bounds, and plug in the definition of $\gamma_m$). 

\begin{lemma}\label{lm:oracle}Consider Setup \ref{setup:2}. For all $m\ge2$, with probability at least $1-{\delta}/(2m^2)$,
we have:
\begin{align*}
\forall t\in\{\tau_{m-2}+1,\cdots,\tau_{m-1}\},~%
\E_{x_t,a_t}\left[(\widehat{f}_{m}(x_t,a_t)-f^*(x_t,a_t))^2\mid\Upsilon_{t-1}\right]%
\le\cE_{\cF,\delta/(2m^2)}(\tau_{m-1}-\tau_{m-2})=\frac{K}{4\gamma_m^2}.
\end{align*}
Therefore (by a union bound), the following event $\Gamma_2$ holds with probability at least $1-\delta/2$:
\begin{equation*}\label{eq:clean-event-inf}
\Gamma_2:=\left\{\forall m\ge 2\text{ and } t\text{ in epoch }m,~~\E_{x_t,a_t}\left[(\widehat{f}_{m}(x_t,a_t)-f^*(x_t,a_t))^2\mid\Upsilon_{t-1}\right]\le \frac{K}{4\gamma_m^2}\right\}.
\end{equation*}
\end{lemma}
\proof{Proof of Lemma \ref{lm:oracle}.} 
Note that Algorithm \ref{alg:falcon+} always sends $(x_t,a_t;r_t(a_t))$-type data to $\texttt{OffReg}_{\cF}$, where $(x_t,r_t)\sim\mcD$ and $a_t\sim p_{m(t)-1}(\cdot\mid x_t)$. %
By Assumption \ref{assump:oracle} and the specification of Algorithm \ref{alg:falcon+}, we have $\forall t\in\{\tau_{m-2}+1,\cdots,\tau_{m-1}\}$,
\begin{align*}
\E_{x_t,a_t}\left[(\widehat{f}_{m}(x_t,a_t)-f^*(x_t,a_t))^2\mid\Upsilon_{t-1}\right]&=\E_{x_t\sim\cD,a_t\sim p_{m-1}(\cdot\mid x_t)}\left[(\widehat{f}_{m}(x_t,a_t)-f^*(x_t,a_t))^2\mid p_{m-1}\right]\\&\le\cE_{\cF,\delta/(2m^2)}(\tau_{m-1}-\tau_{m-2})={K}/({4\gamma_m^2}),
\end{align*}
where the inequality simply follows from Assumption \ref{assump:oracle}.
\Halmos
\endproof

\subsection{Per-Epoch Properties of the Algorithm}\label{app:per-epoch}
We start to utilize the specific properties of our algorithm's action selection kernels to prove our regret bound. We start with some per-epoch properties that always hold for our algorithm regardless of its performance in other epochs. All results in this part hold for both Setup \ref{setup:1} and Setup \ref{setup:2}.

As we mentioned in the main article, a starting point of our proof is to translate the action selection kernel $p_m(\cdot\mid\cdot)$  into an equivalent distribution over policies $Q_m(\cdot)$. Lemma \ref{lm:map} provides a justification of such translation by showing the existence of an equivalent $Q_m$ for every $p_m(\cdot\mid\cdot)$.

\begin{lemma}\label{lm:map}
Fix any epoch $m\in\N$. The action selection scheme $p_m(\cdot\mid\cdot)$ is a valid probability kernel $\cB(\cA)\times\cX\rightarrow[0,1]$. There exists a probability measure $Q_m$ on $\Psi$ such that
$$
\forall a\in\cA,\forall x\in\cX,~~~p_{m}(a\mid x)=\sum_{\pi\in{\Psi}}\1\{\pi(x)=a\}Q_m(\pi). 
$$
\end{lemma}
\proof{Proof of Lemma \ref{lm:map}.}This proof is straightforward when $|\cX|<\infty$. Since  $(\cA,\cB(\cA),p_m(\cdot\mid x))$ is a probability space for each $x\in\cX$, by the existence and uniqueness of finite product probability measures, there exists a unique probability measure $Q_m:=\prod_{x\in\cX}p_m(\cdot\mid x)$ on $(\Psi,(\cB(\cA))^\cX)=(\cA^{\cX},\cB(\cA)^{\cX})$ with the property that 
\[Q_m\left(\prod_{x\in\cX}E_x\right)=\prod_{x\in\cX}p_m(E_x\mid x)\] 
whenever one has $E_x\in\cB(\cA)$ for all $x\in\cX$. For any $a_0\in\cA,x_0\in\cX$, by letting $E_{x_0}=\{a_0\}$ and $E_{x}=\cA$ for all $x\ne x_0$, we have $p_{m}(a_0\mid x_0)=Q_m(\{\pi\,:\,\pi(x_0)=a_0\})=\sum_{\pi\in{\Psi}}\1\{\pi(x_0)=a_0\}Q_m(\pi)$.

\emph{Remark. }In fact, Lemma \ref{lm:map} generally holds for an arbitrary (possibly uncountable) $\cX$, if the equation $p_{m}(a\mid x)=\sum_{\pi\in{\Psi}}\1\{\pi(x)=a\}Q_m(\pi)$ in the statement is replaced by its more general form $p_{m}(a\mid x)=\E_{\pi\sim Q_m}[\1\{\pi(x)=a\}]$. Such a result can be easily obtained by applying the Kolmogorov extension theorem (see, e.g., Theorem 2.4.4 in \citealt{tao2013introduction}).
\Halmos
\endproof

We call the $Q_m$ determined in the proof of Lemma \ref{lm:map} the ``equivalent randomized policy'' induced by $p_m(\cdot\mid\cdot)$. Lemma \ref{lm:equi} states a key property of $Q_m$: the expected instantaneous regret incurred by $p_m(\cdot\mid\cdot)$  equals to the implicit regret of the randomized policy $Q_m$. Thus, to analyze our algorithm's expected regret, we only need to analyze the induced randomized policies' implicit regret.

\begin{lemma}\label{lm:equi}
Fix any epoch $m\in\N$, for any round $t$ in epoch $m$, we have
$$
\E_{x_t,r_t,a_t}\left[r_t(\pi_{f^*})-r_t(a_t)\mid\Upsilon_{t-1}\right]=\sum_{\pi\in\Psi}Q_m(\pi)\Reg(\pi).
$$
\end{lemma}
\proof{Proof of Lemma \ref{lm:equi}.}By Lemma \ref{lm:map}, we have
\begin{align*}
\E_{x_t,r_t,a_t}\left[r_t(\pi_{f^*})-r_t(a_t)\mid\Upsilon_{t-1}\right]&=\E_{x_t,a_t}\left[f^*(x_t,\pi_{f^*}(x_t))-f^*(x_t,a_t)\mid\Upsilon_{t-1}\right]\\
&=\E_{x\sim\cD,a\sim p_m(\cdot\mid x)}\left[f^*(x,\pi_{f^*}(x))-f^*(x,a)\right]\\
&=\E_x\left[\sum_{a\in\cA}p_m(a\mid x)\left(f^*(x,\pi_{f^*}(x))-f^*(x,a)\right)\right]\\
&=\E_x\left[\sum_{a\in\cA}\sum_{\pi\in\Psi}\1\{\pi(x)=a\}Q_m(\pi)\left(f^*(x,\pi_{f^*}(x))-f^*(x,a)\right)\right]\\
&=\E_x\left[\sum_{\pi\in\Psi}Q_m(\pi)\left(f^*(x,\pi_{f^*}(x))-f^*(x,\pi(x))\right)\right]\\
&=\sum_{\pi\in\Psi}Q_m(\pi)\E_x\left[f^*(x,\pi_{f^*}(x))-f^*(x,\pi(x))\right]\\
&=\sum_{\pi\in\Psi}Q_m(\pi)\Reg(\pi).  \Halmos
\end{align*}
\endproof

Lemma \ref{lm:op1} states another key property of $Q_m$. It says that $Q_m$ controls its  predicted implicit regret (relative to the greedy policy based on $\widehat{f}_m$) within $K/\gamma_m$. Note that the controlled error $K/\gamma_m$ is gradually shrinking as the algorithm finishes more epochs.

\begin{lemma}\label{lm:op1}Fix any epoch $m\in\N$, for any round $t$ in epoch $m$, we have
$$
\sum_{\pi\in\Psi}Q_m(\pi)\wReg_{t}(\pi)\le\frac{K}{\gamma_m}.
$$
\end{lemma}
\proof{Proof of Lemma \ref{lm:op1}.}
We have
\begin{align*}
    \sum_{\pi\in\Psi}Q_{m}(\pi)\wReg_{t}(\pi)&=\sum_{\pi\in\Psi}Q_m(\pi)\E_{x\sim\cD}\left[\widehat{f}_{m}(x,\widehat{a}_m(x))-\widehat{f}_m(x,\pi(x))\right]\\
    &=\E_{x\sim\cD}\left[\sum_{\pi\in\Psi}Q_m(\pi)\left(\widehat{f}_{m}(x,\widehat{a}_m(x))-\widehat{f}_m(x,\pi(x))\right)\right]\\
    &=\E_{x\sim\cD}\left[\sum_{a\in\cA}\sum_{\pi\in\Psi}\1\{\pi(x)=a\}Q_m(\pi)\left(\widehat{f}_{m}(x,\widehat{a}_m(x))-\widehat{f}_m(x,a)\right)\right]\\
    &=\E_{x\sim\cD}\left[\sum_{a\in\cA}p_m(a\mid x)\left(\widehat{f}_{m}(x,\widehat{a}_m(x))-\widehat{f}_m(x,a)\right)\right].
\end{align*}
Given any context $x\in\cX$,
$$
\sum_{a\in\cA}p_m(a\mid x)\left(\widehat{f}_{m}(x,\widehat{a}_m(x))-\widehat{f}_m(x,a)\right)=\sum_{a\ne\widehat{a}_m(x)}\frac{\widehat{f}_{m}(x,\widehat{a}_m(x))-\widehat{f}_m(x,a)}{K+\gamma_m\left(\widehat{f}_{m}(x,\widehat{a}_m(x))-\widehat{f}_m(x,a)\right)}\le\frac{K-1}{\gamma_m}.
$$
Lemma \ref{lm:op1} follows immediately.
\Halmos
\endproof

Lemma \ref{lm:op2} states another key per-epoch property of our algorithm. %
For any deterministic policy $\pi\in\Psi$, the quantity $V(p_m,\pi)=\E_{x\sim\cD}\left[{\frac{1}{p_m(\pi(x)\mid x)}}\right]$ is the expected inverse probability that the algorithm's decision generated by $p_m$ (i.e., the decision generated by the randomized policy $Q_m$) is the same as the decision generated by the deterministic policy $\pi$, over the randomization of context $x$. This can be intuitively understood as a  measure of the ``decisional divergence'' between the randomized policy $Q_m$ and the deterministic policy $\pi$.
Lemma \ref{lm:op2} states that this divergence can be bounded by the predicted implicit regret of policy $\pi$.

\begin{lemma}\label{lm:op2}Fix any epoch $m\in\N$, for any round $t$ in epoch $m$, for any policy $\pi\in\Psi$,
$$V(p_m,\pi)\le K+\gamma_m\wReg_{t}(\pi).$$
\end{lemma}
\proof{Proof of Lemma \ref{lm:op2}.}
For any policy $\pi\in\Psi$, given any context $x\in\cX$,
$$
\frac{1}{p_m(\pi(x)\mid x)}\begin{cases}=K+\gamma_m\left(\widehat{f}_{m}(x,\widehat{a}_m(x))-\widehat{f}_m(x,\pi(x))\right),&\text{if }\pi(x)\ne\widehat{a}_m(x);\\
\le \frac{1}{1/K}=K=K+\gamma_m\left(\widehat{f}_{m}(x,\widehat{a}_m(x))-\widehat{f}_m(x,\pi(x))\right),&\text{if }\pi(x)=\widehat{a}_m(x).
\end{cases}
$$
Thus
$$
V(p_m,\pi)=\E_{x\sim\cD}\left[\frac{1}{p_m(\pi(x)\mid x)}\right]\le K+\gamma_m\E_{x\sim\cD}\left[\widehat{f}_{m}(x,\widehat{a}_m(x))-\widehat{f}_m(x,\pi(x))\right]=K+\gamma_m\wReg_t(\pi)
$$
for every round $t$ in epoch $m$.
\Halmos
\endproof

\subsection{Bounding the Prediction Error of Implicit Rewards}\label{app:reward}
Lemma \ref{lm:reward-estimates} relates the prediction error of the implicit reward of any policy $\pi$ at round $t$ to the value of $\cV_t(\pi)$. Recall that $\Gamma_1$ and $\Gamma_2$ are defined in Appendix \ref{app:hpe}.

\begin{lemma}\label{lm:reward-estimates}
Assume $\Gamma_1$ (resp. $\Gamma_2$) holds under Setup \ref{setup:1} (resp., Setup \ref{setup:2}). For any round $t>\tau_1$, for any $\pi\in\Psi$,
$$
|\wcR_t(\pi)-\cR(\pi)|%
\le\frac{\sqrt{\cV_t(\pi)}\sqrt{K}}{2\gamma_{m(t)}}.
$$
\end{lemma}
\proof{Proof of Lemma \ref{lm:reward-estimates}.}
Fix any policy $\pi\in\Psi$, and any round $t>\tau_1$. 
By the definitions of $\wcR_t(\pi)$ and $\cR(\pi)$, we have
$$\wcR_t(\pi)-\cR(\pi)=\E_{x\sim\cD}\left[\widehat{f}_{m(t)}(x,\pi(x))-f^*(x,\pi(x))\right].$$
Given a context $x$, define
$$
\Delta_x=\widehat{f}_{m(t)}(x,\pi(x))-f^*(x,\pi(x)),
$$
then $\wcR_t(\pi)-\cR(\pi)=\E_{x\sim\cD}[\Delta_x]$.
For all $s=1,2,\dots,\tau_{m(t)-1}$, we have
\begin{align}
\E_{a_s|x_s}\left[\left(\widehat{f}_{m(t)}(x_s,a_s)-f^*(x_s,a_s)\right)^2\mid\Upsilon_{s-1}\right]&=\sum_{a\in\cA}p_{m(s)}(a\mid x_s)\left(\widehat{f}_{m(t)}(x_s,a)-f^*(x_s,a)\right)^2\notag\\&\ge p_{m(s)}(\pi(x_s)\mid x_s)\left(\widehat{f}_{m(t)}(x_s,\pi(x_s))-f^*(x_s,\pi(x_s))\right)^2\notag\\&=p_{m(s)}(\pi(x_s)\mid x_s)\left(\Delta_{x_s}\right)^2.\label{eq:reward-estimation}
\end{align}
Thus for both $s_0=1$ and $s_0=\tau_{m(t)-2}+1$, we have
\begin{align*}
&~~~\cV_t(\pi)\sum_{s=s_0}^{\tau_{m(t)-1}}\E_{x_s,a_s}\left[(\widehat{f}_{m(t)}(x_s,a_s)-f^*(x_s,a_s))^2\mid\Upsilon_{s-1}\right]\\
&\overset{\rm(i)}{\ge}\sum_{s=s_0}^{\tau_{m(t)-1}}V(p_{m(s)},\pi)\E_{x_s,a_s}\left[(\widehat{f}_{m(t)}(x_s,a_s)-f^*(x_s,a_s))^2\mid\Upsilon_{s-1}\right]\\
&=\sum_{s=s_0}^{\tau_{m(t)-1}}\E_{x_s}\left[\frac{1}{p_{m(s)}(\pi(x_s)\mid x_s)}\right]\E_{x_s}\E_{a_s|x_s}\left[\left(\widehat{f}_{m(t)}(x_s,a_s)-f^*(x_s,a_s)\right)^2\mid\Upsilon_{s-1}\right]\\
&\overset{\rm(ii)}{\ge}\sum_{s=s_0}^{\tau_{m(t)-1}}\left(\E_{x_s}\left[\sqrt{\frac{1}{p_{m(s)}(\pi(x_s)\mid x_s)}\E_{a_s|x_s}\left[\left(\widehat{f}_{m(t)}(x_s,a_s)-f^*(x_s,a_s)\right)^2\mid\Upsilon_{s-1}\right]}\right]\right)^2\\
&\overset{\rm(iii)}{\ge}\sum_{s=s_0}^{\tau_{m(t)-1}}\left(\E_{x_s}\left[\sqrt{\frac{1}{p_{m(s)}(\pi(x_s)\mid x_s)}p_{m(s)}(\pi(x_s)\mid x_s)\left(\Delta_{x_s}\right)^2}~\right]\right)^2\\
&=\sum_{s=s_0}^{\tau_{m(t)-1}}\left(\E_{x_s}\left[|\Delta_{x_s}|\right]\right)^2\\
&\overset{\rm(iv)}{\ge}\sum_{s=s_0}^{\tau_{m(t)-1}}|\wcR_t(\pi)-\cR(\pi)|^2=(\tau_{m(t)-1}-s_0+1)|\wcR_t(\pi)-\cR(\pi)|^2,
\end{align*}
where (i) follows from the definition of $\cV_t(\pi)$, (ii) follows from the Cauchy-Schwarz inequality, (iii) follows from (\ref{eq:reward-estimation}), and (iv) follows from the convexity of the $\ell_1$ norm.

If we are in Setup \ref{setup:1} and $\Gamma_1$ holds, then by letting $s_0=1$, we have
\[
|\wcR_t(\pi)-\cR(\pi)|\le \sqrt{\cV_t(\pi)}\sqrt{\frac{\sum_{s=1}^{\tau_{m(t)-1}}\E_{x_s,a_s}\left[(\widehat{f}_{m(t)}(x_s,a_s)-f^*(x_s,a_s))^2\mid\Upsilon_{s-1}\right]}{\tau_{m(t)-1}}}\le\frac{\sqrt{\cV_t(\pi)}\sqrt{K}}{2\gamma_{m(t)}},
\]
where the last inequality follows from the definition of $\Gamma_1$. If we are in Setup \ref{setup:2} and $\Gamma_2$ holds, then by letting $s_0=\tau_{m(t)-2}+1$, we have
\[
|\wcR_t(\pi)-\cR(\pi)|\le \sqrt{\cV_t(\pi)}\sqrt{\frac{\sum_{s=\tau_{m(t)-2}+1}^{\tau_{m(t)-1}}\E_{x_s,a_s}\left[(\widehat{f}_{m(t)}(x_s,a_s)-f^*(x_s,a_s))^2\mid\Upsilon_{s-1}\right]}{\tau_{m(t)-1}-\tau_{m(t)-2}}}\le\frac{\sqrt{\cV_t(\pi)}\sqrt{K}}{2\gamma_{m(t)}},
\]
where the last inequality follows from the definition of $\Gamma_2$.
\Halmos
\endproof

\subsection{Bounding the Prediction Error of Implicit Regret}\label{app:regret}

Lemma \ref{lm:regret-estimates} establishes an important relationship between the predicted implicit regret and the true implicit regret of any policy $\pi$ at round $t$. This lemma ensures that the predicted implicit regret of ``good policies'' are becoming more and more accurate, while the predicted implicit regret of ``bad policies'' do not need to have such property.

Recall that $\Gamma_1$ and $\Gamma_2$ are defined in Appendix \ref{app:hpe}.

\begin{lemma}\label{lm:regret-estimates}
Assume that $\Gamma_1$ (resp. $\Gamma_2$) holds under Setup \ref{setup:1} (resp. Setup \ref{setup:2}). Let $c_0:=5.15$. For all epochs $m\in\N$, all rounds $t$ in epoch $m$, and all policies $\pi\in\Psi$,
$$
\Reg(\pi)\le 2\wReg_t(\pi)+c_0K/\gamma_{m},
$$
$$
\wReg_t(\pi)\le 2\Reg(\pi)+c_0K/\gamma_{m}.
$$
\end{lemma}
\proof{Proof of Lemma \ref{lm:regret-estimates}.}
We prove Lemma \ref{lm:regret-estimates} via induction on $m$. We first consider the base case where $m=1$ and $1\le t\le \tau_1$. In this case, since $\gamma_1=1$, we know that $\forall\pi\in\Psi$, 
\begin{equation*}
\Reg(\pi)\le 1\le c_0 K/\gamma_1,~~~\wReg_t(\pi)\le1\le c_0K/\gamma_1;\tag{\text{under Setup \ref{setup:1}}}
\end{equation*}
\begin{equation*}
\Reg(\pi)\le \sqrt{K}\le c_0 K/\gamma_1,~~~\wReg_t(\pi)=0\le c_0K/\gamma_1.\tag{\text{under Setup \ref{setup:2}}}
\end{equation*}
Note that we use condition (\ref{eq:condition}) for Setup \ref{setup:2} here. Thus the claim holds in the base case.

For the inductive step, fix some epoch $m>1$. We assume that for all epochs $m'<m$, all rounds $t'$ in epoch $m'$, and all $\pi\in\Psi$,
\begin{equation}\label{eq:ind1}
\Reg(\pi)\le 2\wReg_{t'}(\pi)+c_0K/\gamma_{m'},
\end{equation}
\begin{equation}\label{eq:ind2}
\wReg_{t'}(\pi)\le 2\Reg(\pi)+c_0K/\gamma_{m'}.
\end{equation}

We first show that for all rounds $t$ in epoch $m$ and all $\pi\in\Psi$,
$$
\Reg(\pi)\le 2\wReg_{t}(\pi)+c_0K/\gamma_{m}.
$$
We have
\begin{align}\label{eq:ind-dif}
\Reg(\pi)-\wReg_t(\pi)&=(\cR(\pi_{f^*})-\cR(\pi))-(\wcR_t(\pi_{\widehat{f}_{m}})-\wcR_t(\pi))\notag\\
&\le(\cR(\pi_{f^*})-\cR(\pi))-(\wcR_t(\pi_{f^*})-\wcR_t(\pi))\notag\\
&\le |\wcR_t(\pi)-\cR(\pi)|+|\wcR_t(\pi_{f^*})-\cR(\pi_{f^*})|\notag\\
&\le {\frac{\sqrt{\cV_t(\pi)}\sqrt{K}}{2\gamma_{m}}}+\frac{\sqrt{\cV_t(\pi_{f^*})}\sqrt{K}}{2\gamma_{m}}\notag\\
&\le\frac{\cV_t(\pi)}{5\gamma_{m}}+\frac{\cV_t(\pi_{f^*})}{5\gamma_{m}}+\frac{5K}{8\gamma_{m}}
\end{align}
where the first inequality is by the optimality of $\pi_{\widehat{f}_{m}}$ for $\wcR_t(\cdot)$, the second inequality is by the triangle inequality, the third inequality is by Lemma \ref{lm:reward-estimates}, and the fourth inequality is by the AM-GM inequality.  
By the definitions of $\cV_t(\pi),\cV_t(\pi_{f^*})$ and Lemma \ref{lm:op2}, there exist epochs $i,j< m$ such that
$$
\cV_t(\pi)=V(p_i,\pi)=\E_{x\sim\cD}\left[\frac{1}{p_i(\pi(x)\mid x)}\right]\le K+\gamma_{i}\wReg_{\tau_{i}}(\pi),
$$
$$
\cV_t(\pi_{f^*})=V(p_j,\pi_{f^*})=\E_{x\sim\cD}\left[\frac{1}{p_j(\pi_{f^*}(x)\mid x)}\right]\le K+\gamma_{j}\wReg_{\tau_j}(\pi_{f^*}).
$$
Combining the above two inequalities with (\ref{eq:ind2}), we have
\begin{align}\label{eq:ind3}
\frac{\cV_t(\pi)}{5\gamma_{m}}\le \frac{K+\gamma_i\wReg_{\tau_i}(\pi)}{5\gamma_{m}}\le \frac{K+\gamma_{i}(2\Reg(\pi)+c_0K/\gamma_{i})}{5\gamma_{m}}\le\frac{(1+c_0)K}{5\gamma_{m}}+\frac{2}{5}\Reg(\pi),
\end{align}
\begin{align}\label{eq:ind4}
\frac{\cV_t(\pi_{f^*})}{5\gamma_{m}}\le \frac{K+\gamma_j\wReg_{\tau_j}(\pi_{f^*})}{5\gamma_{m}}\le \frac{K+\gamma_{j}(2\Reg(\pi_{f^*})+c_0K/\gamma_{j})}{5\gamma_{m}}=\frac{(1+c_0)K}{5\gamma_{m}},
\end{align}
where the last inequality in (\ref{eq:ind3}) follows from $\gamma_i\le\gamma_m$ and the last inequality in (\ref{eq:ind4}) follows from $\Reg(\pi_{f^*})=0$. Combining (\ref{eq:ind-dif}), (\ref{eq:ind3}) and (\ref{eq:ind4}), we have
\begin{equation}\label{eq:ind5}
\Reg(\pi)\le\frac{5}{3}\wReg_t(\pi)+\frac{2c_0K}{3\gamma_{m}}+\frac{1.71K}{\gamma_{m}}\le 2\wReg_t(\pi)+\frac{c_0K}{\gamma_{m}}.
\end{equation}

We then show that for all rounds $t$ in epoch $m$ and all $\pi\in\Psi$,
$$
\wReg_t(\pi)\le 2\Reg_{t}(\pi)+c_0K/\gamma_{m}.
$$
Similar to (\ref{eq:ind-dif}), we have
\begin{align}\label{eq:ind-dif'}
\wReg_t(\pi)-\Reg(\pi)&=(\wcR(\pi_{\widehat{f}_{m}})-\wcR_t(\pi))-(\cR(\pi_{f^*})-\cR(\pi))\notag\\
&\le(\wcR_t(\pi_{\widehat{f}_{m}})-\wcR_t(\pi))-(\cR(\pi_{{\widehat{f}_{m}}})-\cR(\pi))\notag\\
&\le |\wcR_t(\pi)-\cR(\pi)|+|\wcR_t(\pi_{\widehat{f}_{m}})-\cR(\pi_{\widehat{f}_{m}})|\notag\\
&\le {\frac{\sqrt{\cV_t(\pi)}\sqrt{K}}{\gamma_{m}}}+\frac{\sqrt{\cV_t(\pi_{\widehat{f}_{m}})}\sqrt{K}}{\gamma_{m}}\notag\\
&\le\frac{\cV_t(\pi)}{5\gamma_{m}}+\frac{\cV_t(\pi_{\widehat{f}_{m}})}{5\gamma_{m}}+\frac{5K}{8\gamma_{m}}.
\end{align}
By the definition of $\cV_t(\pi_{\widehat{f}_m})$ and Lemma \ref{lm:op2}, there exist epoch $l< m$ such that
$$
\cV_t(\pi_{\widehat{f}_m})=V(p_l,\pi_{\widehat{f}_m})=\E_{x\sim\cD}\left[\frac{1}{p_l(\pi_{\widehat{f}_m}|x)}\right]\le K+\gamma_{l}\wReg_{\tau_{l}}(\pi_{\widehat{f}_m}).
$$
Using (\ref{eq:ind2}), $\gamma_l\le\gamma_{m}$, (\ref{eq:ind5}) and $\wReg_{t}(\pi_{\widehat{f}_m})=0$, we have
\begin{align}\label{eq:ind6}
\frac{\cV_t(\pi_{\widehat{f}_m})}{5\gamma_{m}}&\le \frac{K+\gamma_l\wReg_{\tau_l}(\pi_{\widehat{f}_m})}{5\gamma_{m}}\le \frac{K+\gamma_{l}(2\Reg(\pi_{\widehat{f}_m})+c_0K/\gamma_{l})}{5\gamma_{m}}\le\frac{(1+c_0)K}{5\gamma_{m}}+\frac{2}{5}\Reg(\pi_{\widehat{f}_m})\notag\\
&\le\frac{(1+c_0)K}{5\gamma_{m}}+\frac{2}{5}\left(\wReg_t(\pi_{\widehat{f}_m})+\frac{c_0K}{\gamma_{m}}\right)=\frac{(1+3c_0)K}{5\gamma_{m}}.
\end{align}
Combining (\ref{eq:ind3}), (\ref{eq:ind-dif'}) and (\ref{eq:ind6}), we have
\begin{align*}
    \wReg_t(\pi)\le\frac{7}{5}\Reg(\pi)+\frac{4c_0K}{5\gamma_{m}}+\frac{1.03K}{\gamma_{m}}\le2\Reg(\pi)+\frac{c_0K}{\gamma_{m}}.
\end{align*}
Thus we complete the inductive step, and the claim proves to be true for all $m\in\N$.
\Halmos

\endproof

\subsection{Bounding the True Regret}\label{app:final}
In this part, we  put everything together and finally prove Lemma \ref{lm:final}, which holds for both Setups \ref{setup:1} and \ref{setup:2}, and simultaneously implies Theorem \ref{thm:finite}, Corollary \ref{cor:finite}, and Theorem \ref{thm:inf}.  Moreover, Lemma \ref{lm:final} implies that  bounded rewards are not required if we only want to bound the expected regret.

\begin{lemma}\label{lm:true-reg}Recall that $\Gamma_1$ and $\Gamma_2$ are defined in Appendix \ref{app:hpe}. 
Assume that $\Gamma_1$ (resp. $\Gamma_2$) holds under Setup \ref{setup:1} (resp. Setup \ref{setup:2}). For every epoch $m\in\N$,
$$
\sum_{\pi\in\Psi}Q_m(\pi)\Reg(\pi)\le7.15K/\gamma_m.
$$
\end{lemma}
\proof{Proof of Lemma \ref{lm:true-reg}.}
Fix any epoch $m\in\N$. Since $\tau_{m-1}+1$ belongs to epoch $m$, we have
\begin{align*}
    \sum_{\pi\in\Psi}Q_m(\pi)\Reg(\pi)&\le\sum_{\pi\in\Psi}Q_m(\pi)\left(2{\wReg}_{\tau_{m-1}+1}(\pi)+\frac{c_0K}{\gamma_{m}}\right)\\
    &=2\sum_{\pi\in\Psi}Q_m(\pi){\wReg}_{\tau_{m-1}+1}(\pi)+\frac{c_0K}{\gamma_m}\\
    &\le\frac{(2+c_0)K}{\gamma_m},
\end{align*}
where the first inequality follows from Lemma \ref{lm:regret-estimates}, and the second inequality follows from Lemma \ref{lm:op1}. We then take in $c_0=5.15$. 
\Halmos

\endproof

\begin{lemma}\label{lm:final}
For any  $T\in\N$, the expected regret of our algorithm after $T$ rounds is at most $\sum_{t=\tau_1+1}^{T}7.15K/\gamma_{m(t)}+\sqrt{K}\tau_1+T\delta/2$.
Furthermore, if all rewards are $[0,1]$-bounded, then with probability at least $1-\delta$, the regret after $T$ rounds is at most
\[
\sum_{t=\tau_1+1}^{T}7.15K/\gamma_{m(t)}+\tau_1+\sqrt{8T\log(2/\delta)}.
\]
\end{lemma}
\proof{Proof of Lemma \ref{lm:final}.}
Fix $T\in\N$. Since $\Gamma_1$ (resp. $\Gamma_2$) holds under Setup \ref{setup:1} (resp. Setup \ref{setup:2}) with probability at least $1-\delta/2$, by Lemma \ref{lm:equi} and Lemma \ref{lm:true-reg}, we can bound the expected regret:
\begin{align*}
&\E\left[\sum_{t=1}^T\left(r_t(\pi_{f^*})-r_t(a_t)\right)\right]=\E\left[\sum_{t=1}^T\E_{x_t,r_t,a_t}\left[r_t(\pi_{f^*})-r_t(a_t)\mid\Upsilon_{t-1}\right]\right]\\&=\E\left[\sum_{t=1}^T\sum_{\pi\in\Psi}Q_{m(t)}(\pi)\Reg(\pi)\right]\le\sum_{t=\tau_1+1}^{T}7.15K/\gamma_{m(t)}+\sqrt{K}\tau_1+T\delta/2.
\end{align*}
We now assume $r_t\in[0,1]^K$ and turn to the high-probability bound. 
For each round $t\in\{1,\dots,T\}$, define  $M_t:=r_t(\pi_{f^*})-r_t(a_t)-\sum_{\pi\in\Psi}Q_{m(t)}(\pi)\Reg(\pi)$. By Lemma \ref{lm:equi} we have
$$
\E_{x_t,r_t,a_t}\left[r_t(\pi_{f^*})-r_t(a_t)\mid\Upsilon_{t-1}\right]=\sum_{\pi\in\Psi}Q_{m(t)}(\pi)\Reg(\pi),~~~\E_{x_t,r_t,a_t}[M_t\mid\Upsilon_{t-1}]=0,
$$
Since $|M_t|\le 2$, $M_t$ is a martingale difference sequence. By Azuma's inequality,
\begin{equation}\label{eq:net}
\sum_{t=1}^T M_t\le2\sqrt{2T\log(2/\delta)}
\end{equation}
with probability at least $1-{\delta}/{2}$. By Lemma \ref{lm:ag12-2} (resp. Lemma \ref{lm:oracle}), with probability at least $1-\delta/2$, the event $\Gamma_1$ (resp. $\Gamma_2$) holds. By a union bound, with probability at least $1-\delta$, %
\begin{align*}
\sum_{t=1}^T\left(r_t(\pi_{f^*})-r_t(a_t)\right)&{\le}\sum_{t=1}^T\sum_{\pi\in\Psi}Q_{m(t)}(\pi)\Reg(\pi)+\sqrt{8T\log(2/\delta)}\\
&{\le}\sum_{t=\tau_1+1}^{T}7.15K/\gamma_{m(t)}+\tau_1+\sqrt{8T\log(2/\delta)}
\end{align*}
where the first inequality follows from (\ref{eq:net}), and the second inequality follows from Lemma \ref{lm:true-reg}.
\Halmos
\endproof

Finally, we assume that all rewards are $[0,1]$-bounded and use Lemma \ref{lm:final} to derive Theorem \ref{thm:finite}, Corollary \ref{cor:finite}, and Theorem \ref{thm:inf}.

\proof{Proof of Theorem \ref{thm:finite}.}
We are in Setup \ref{setup:1}, and we have $\tau_m\le 2^m,\,\forall m\in\N$ and $\tau_m\le2\tau_{m-1},\,\forall m>1$. By Lemma \ref{lm:final}, with probability at least $1-\delta$,
\begin{align*}
\sum_{t=1}^T\left(r_t(\pi_{f^*})-r_t(a_t)\right)&{\le}\sum_{t=\tau_1+1}^{T}7.15K/\gamma_{m(t)}+\tau_1+\sqrt{8T\log(2/\delta)}\\
&=215\sum_{m=2}^{m(T)}\sqrt{\frac{K\log(|\cF|\log(\tau_{m-1})m/\delta)}{\tau_{m-1}}}(\tau_m-\tau_{m-1})+\tau_1+\sqrt{8T\log(2/\delta)}\\
&\overset{\rm(i)}{\le} 215\sqrt{K\log(|\cF|m(T)^2/\delta)}\sum_{m=2}^{m(T)}\frac{\tau_m-\tau_{m-1}}{\sqrt{\tau_{m}/2}}+\sqrt{8T\log(2/\delta)}+\tau_1\\
&\overset{\rm(ii)}{\le}215\sqrt{2K\log(|\cF|m(T)^2/\delta)}\sum_{m=2}^{m(T)}\int_{\tau_{m-1}}^{\tau_m}\frac{dx}{\sqrt{x}}+\sqrt{8T\log(2/\delta)}+\tau_1\\
&=215\sqrt{2K\log(|\cF|m(T)^2/\delta)}\int_{\tau_{1}}^{\tau_{m(T)}}\frac{dx}{\sqrt{x}}+\sqrt{8T\log(2/\delta)}+\tau_1\\
&\le430\sqrt{2\tau_{m(T
)}K\log(|\cF|m(T)^2/\delta)}+\sqrt{8T\log(2/\delta)}+\tau_1\\
&\overset{\rm(iii)}{\le}860 \sqrt{KT\log(|\cF|m(T)^2/\delta)}+\sqrt{8T\log(2/\delta)}+\tau_1,
\end{align*}
where (i) follows from $\log(\tau_{m-1})\le m-1 \le m(T)$ and $\tau_m\le 2\tau_{m-1}$, (ii) follows from an integral bound, and (iii) follows from $\tau_{m(T)}\le2\tau_{m(T)-1}<2T$.  We thus finish our proof of Theorem \ref{thm:finite}.\Halmos
\endproof

\proof{Proof of Corollary \ref{cor:finite}.} We are in Setup \ref{setup:1}, and we have  $\tau_m=\left\lfloor 2T^{1-2^{-m}}\right\rfloor$, $\forall m\in\N$. Without loss of generality, assume that $T>1000$.%

By Lemma \ref{lm:final}, with probability at least $1-\delta$, we have
\begin{align*}
\sum_{t=1}^T\left(r_t(\pi_{f^*})-r_t(a_t)\right)
&\le\sum_{t=\tau_1+1}^{T}7.15K/\gamma_{m(t)}+\tau_1+\sqrt{8T\log(2/\delta)}\\
&=215\sum_{m=2}^{m(T)}\sqrt{\frac{K\log(|\cF|m\log(\tau_{m-1})/\delta)}{\tau_{m-1}}}(\tau_m-\tau_{m-1})+\tau_1+\sqrt{8T\log(2/\delta)}\\
&\le215\sqrt{K\log(|\cF|m(T)\log T/\delta)}\sum_{m=2}^{m(T)}\frac{\tau_m-\tau_{m-1}}{\sqrt{\tau_{m-1}}}+\sqrt{8T\log(2/\delta)}+\tau_1\\
&\le 215\sqrt{K\log(|\cF|m(T)\log T/\delta)}\sum_{m=2}^{m(T)}\frac{\tau_m}{\sqrt{\tau_{m-1}}}+\sqrt{8T\log(2/\delta)}+\tau_1\\
&\le 215\sqrt{K\log(|\cF|m(T)\log T/\delta)}\left(2\sqrt{T}\right)(m(T)-1)+\sqrt{8T\log(2/\delta)}+2\sqrt{T},
\end{align*}
where the last inequality follows from
$$
\frac{\tau_m}{\sqrt{\tau_{m-1}}}\le \frac{\tau_m}{\sqrt{(\tau_{m-1}+1)/2}}\le\frac{2T^{1-2^{-m}}}{T^{\frac{1}{2}(1-2^{-m+1})}}=2\sqrt{T},~~~\forall m>1
$$
and $\tau_1\le2\sqrt{T}$. Corollary \ref{cor:finite} follows from the fact that $m(T)=O(\log\log T)$.
\Halmos
\endproof

\proof{Proof of Theorem \ref{thm:inf}.} We are in Setup \ref{setup:2}, and we have assumed that all rewards are $[0,1]$-bounded.

By Lemma \ref{lm:final}, with probability at least $1-\delta$, we have
\begin{align*}
\sum_{t=1}^T\left(r_t(\pi_{f^*})-r_t(a_t)\right)%
&\le\sum_{t=\tau_1+1}^{T}7.15K/\gamma_{m(t)}+\tau_1+\sqrt{8T\log(2/\delta)}\\
&=14.3\sum_{m=2}^{m(T)}\sqrt{K\cE_{\cF,\delta/(2m^2)}(\tau_{m-1}-\tau_{m-2})}(\tau_m-\tau_{m-1})+\tau_1+\sqrt{8T\log(2/\delta)},%
\end{align*}
where we directly plug in the definition of $\gamma_m$. 
\Halmos
\endproof

\emph{Remark. }Note that the entire proof of Theorem \ref{thm:inf} holds without assuming $f^*\in\cF$. Therefore, the extension to the misspecified setting  described in \S\ref{sec:advantages} (where the misspecification error is known) is straightforward. %

\subsection{Dealing with Uncountable Context Spaces}\label{app:measure}

We have proved Theorem \ref{thm:finite} and Theorem \ref{thm:inf} under the condition that $|\cX|<\infty$ (note that we allow $|\cX|$ to be arbitrarily large in this setting, as the regret bound does not depend on $|\cX|$). The main role of the condition $|\cX|<\infty$ is that it makes all policies of the form $\pi:\cX\rightarrow\cA$ automatically measurable with respect to $\cD$, enabling us to define $\cR(\cdot),\wcR_t(\cdot),\Reg(\cdot),\wReg_t(\cdot)$ for all policies $\pi\in\cA^{\cX}$ without worrying about any measurability issues. Since everything is well-defined in the universal policy space, we are able to provide an illuminating analysis in this space, which not only explains the value and role of realizability (or relaxed notions of realizability) but also establishes new connections among several research lines of contextual bandits.

Nevertheless, to ensure that Theorem \ref{thm:finite} and Theorem \ref{thm:inf} indeed hold for a generic, possibly uncountable $\cX$, we need to deal with the potential measurability issues associated with $\cA^{\cX}$ when $\cX$ is arbitrary. In what follows, we discuss such issues and give a simple resolution.%

Before we proceed, we make two remarks. First, to ensure that the contextual bandit problem that we study is meaningful, some basic conditions are required, e.g., 
the true reward function $f^*$ should be measurable with respect to $\cD$ for each fixed $a\in\cA$, and the regression oracle should always generate predictors with such a property. Such conditions are necessary for the regret of the algorithm to be well-defined, and are directly assumed here.

Second, in terms of our analysis, the only issue that is worthy of special attention when $\cX$ is uncountable is that, not all policies in $\Psi=\cA^{\cX}$ are guaranteed to be measurable with respect to $\cD$ (as a consequence,  $\cR(\cdot),\wcR_t(\cdot),\Reg(\cdot),\wReg_t(\cdot)$ may not be ``everywhere defined'' on $\Psi$). %
All other issues, such as the existence and properties of $Q_m(\cdot)$, can be easily addressed by standard tools from measure theory (e.g., Theorem 2.4.4 of \citealt{tao2013introduction}). %

We now focus on the key issue that $\Psi$ may contain non-measurable policies (when $\cX$ is uncountable). It turns out that the affect of this issue on our previous analysis is mostly ``notational.'' Namely, since $\cR(\cdot),\wcR_t(\cdot),\Reg(\cdot),\wReg_t(\cdot)$ are not necessarily well-defined for all $\pi\in\Psi$,  Lemma \ref{lm:reward-estimates} and Lemma \ref{lm:regret-estimates}---which involves the universal quantifier  ``for all $\pi\in\Psi$''---require slight modifications.

There are multiple ways to address such an issue. We provide a a simple resolution below, which works for general uncountable $\cX$ and  generates additional insights about our proof. 

The resolution is based on the following observation: while $\cR(\pi)=\E_{x\sim\cD}\left[f^*(x,\pi(x))\right]$ is not guaranteed to be well-defined for an arbitrary deterministic policy $\pi\in\Psi$, the quantity $\cR(Q_m)=\E_{x\sim\cD}\E_{\pi\sim Q_m}\left[f^*(x,\pi(x))\right]$ is always well-defined, as the algorithm's adopted randomized policy $Q_m(\cdot)$ in epoch $m$ is ``measurable.'' Thus,
we can directly define $\cR(Q)=\E_{x\sim\cD}\E_{\pi\sim Q}\left[f^*(x,\pi(x))\right]$ for all ``measurable'' randomized policies $Q(\cdot)$, and similarly define $\wcR_t(Q),\Reg(Q),\wReg_t(Q)$. This would enable us to restate Lemma \ref{lm:reward-estimates} and Lemma \ref{lm:regret-estimates} by replacing ``for all deterministic policies $\pi\in\Psi$'' with ``for all measurable randomized policies $Q$,'' and there would be no measurability issues any more. Moreover, observing that an action selection kernel $p:\cB(\cA)\times\cX\rightarrow[0,1]$ exactly corresponds to a ``measurable'' randomized policy (as the definition of a probability kernel clearly requires $p(E\mid x)$ to be measurable with respect to $\cD$ for any fixed $E\in\cB(\cA)$), we can actually directly use the kernel $p(\cdot\mid\cdot)$ to denote a ``measurable'' randomized policy, which would lead to a new version of our proof without explicit use of the notation $Q_m(\cdot)$ or $Q(\cdot)$. 

Specifically, the new general proof proceeds as follows. Recall that $\cP$ is the space of all action selection kernels, which corresponds to the space of all \emph{measurable randomized policies}. 
Analogous to Appendix \ref{app:def}, for all (measurable) randomized policies $p,p'\in\cP$ and $t=1,\dots,T$, define
\[
V(p,p'):=\E_{x\sim\cD,a\sim p'(\cdot\mid x)}\left[\frac{1}{p{(a\mid x)}}\right],~~~~
\cV_t(p):=\max_{1\le m \le m(t)-1}\{V(p_m,p)\},
\]
\[
\cR(p):=\E_{x\sim\cD,a\sim p(\cdot\mid x)}\left[f^*(x,a)\right],~~~~
\wcR_t(p):=\E_{x\sim\cD,a\sim p(\cdot\mid x)}\left[\widehat{f}_{m(t)}(x,a)\right],
\]
\[
\Reg(p):=\cR(\pi_{f^*})-\cR(p),~~~~
\wReg_t(p):=\wcR_t(\pi_{\widehat{f}_{m(t)}})-\wcR_t(p),
\]
where $V(p,p')$ is the ``decisional divergence'' between two randomized policies $p$ and $p'$. Analogous to Appendix \ref{app:per-epoch}, we can show that in each epoch $m\in\N$, the algorithm's adopted randomized policy $p_m\in\cP$ is a solution to the following ``Implicit Optimization Problem'':
\begin{align*}
    \wReg_{t}(p_m)&\le {K}/{\gamma_m},\\
    \forall p\in\cP,~~~\E_{x\sim\cD,a\sim p(\cdot\mid x)}\left[\frac{1}{p_m{(a\mid x)}}\right]&\le K+\gamma_m\wReg_{t}(p).
\end{align*}
Analogous to Lemma \ref{lm:reward-estimates} to Lemma \ref{lm:regret-estimates}, we have the following guarantees. 
\begin{lemma}\label{lm:reward-estimates-new}
Assume $\Gamma_1$ (resp. $\Gamma_2$) holds under Setup \ref{setup:1} (resp., Setup \ref{setup:2}). For all rounds $t>\tau_1$, for all (measurable) randomized policies $p\in\cP$,
$$
|\wcR_t(p)-\cR(p)|%
\le\frac{\sqrt{\cV_t(p)}\sqrt{K}}{2\gamma_{m(t)}}.
$$
\end{lemma}

\begin{lemma}\label{lm:regret-estimates-new}
Assume that $\Gamma_1$ (resp. $\Gamma_2$) holds under Setup \ref{setup:1} (resp. Setup \ref{setup:2}). Let $c_0:=5.15$. For all epochs $m\in\N$, all rounds $t$ in epoch $m$, and all (measurable) randomized policies $p\in\cP$,
$$
\Reg(p)\le 2\wReg_t(p)+c_0K/\gamma_{m},
$$
$$
\wReg_t(p)\le 2\Reg(p)+c_0K/\gamma_{m}.
$$
\end{lemma}

Lemma \ref{lm:reward-estimates-new} and Lemma \ref{lm:regret-estimates-new} are almost identical to  Lemma \ref{lm:reward-estimates} and Lemma \ref{lm:regret-estimates}, except for the slight difference that ``for all $\pi\in\Psi$'' is replaced with ``for all $p\in\cP$,'' which ensures that there are no measurability issues (as $\cR(\cdot),\wcR_t(\cdot),\Reg(\cdot),\wReg_t(\cdot)$ are ``everywhere defined'' on $\cP$). We then easily have $\E_{x_t,r_t,a_t}\left[r_t(\pi_{f^*})-r_t(a_t)\mid\Upsilon_{t-1}\right]=\Reg(p_{m(t)})\le7.15K/\gamma_{m(t)}$, and Theorem \ref{thm:finite} and Theorem \ref{thm:inf} immediately follow. The above proof is essentially the same as the proof that we provide in  Appendix \ref{app:def} to Appendix \ref{app:final}, but under a different set of notations. %

\end{appendices}

\end{document}